\documentclass[10pt,twocolumn,letterpaper]{article}

\usepackage{amsmath}
\usepackage{amssymb}
\usepackage{amsthm}
\usepackage[ruled, lined, linesnumbered, commentsnumbered, longend]{algorithm2e}
\usepackage{array}
\usepackage[caption=false,font=normalsize,labelfont=sf,textfont=sf]{subfig}
\usepackage{textcomp}
\usepackage{stfloats}
\usepackage{verbatim}
\usepackage{graphicx}
\usepackage{cite}
\usepackage{xcolor}

\usepackage{booktabs}
\usepackage{multirow}

\DeclareMathOperator*{\argmin}{arg\,min} 
\DeclareMathOperator*{\argmax}{arg\,max}

\usepackage{booktabs, siunitx, multirow, adjustbox, makecell}
\newcommand{\mypara}[1]{\vspace{0.2cm}\noindent\textbf{#1.}~}

\usepackage{cvpr}              

\definecolor{cvprblue}{rgb}{0.21,0.49,0.74}
\usepackage[pagebackref,breaklinks,colorlinks,allcolors=cvprblue]{hyperref}

\def\paperID{*****} %
\def\confName{CVPR}
\def\confYear{2026}

\title{DualReg: Dual-Space Filtering and Reinforcement for Rigid Registration}

\author{
Jiayi Li$^{1}$, 
Yuxin Yao$^{2}$\footnotemark[1]~, 
Qiuhang Lu$^{3}$, 
Juyong Zhang$^{1}$\thanks{Corresponding authors.} \\
$^{1}$University of Science and Technology of China, \\
$^{2}$City University of Hong Kong, 
$^{3}$University of Chinese Academy of Sciences\\
{\tt\small SA23001023@mail.ustc.edu.cn;yuxinyao@cityu.edu.hk;luqiuhang2021@ia.ac.cn;juyong@ustc.edu.cn}
}

\begin{document}
\maketitle
\begin{abstract}
Noisy, partially overlapping data and the need for real-time processing pose major challenges for rigid registration. Considering that feature-based matching can handle large transformation differences but suffers from limited accuracy, while local geometry-based matching can achieve fine-grained local alignment but relies heavily on a good initial transformation, we propose a novel dual-space paradigm to fully leverage the strengths of both approaches. First, we introduce an efficient filtering mechanism consisting of a computationally lightweight one-point RANSAC algorithm and a subsequent refinement module to eliminate unreliable feature-based correspondences. Subsequently, we treat the filtered correspondences as anchor points, extract geometric proxies, and formulate an effective objective function with a tailored solver to estimate the transformation. Experiments verify our method's effectiveness, as demonstrated by a 32x CPU-time speedup over MAC on KITTI with comparable accuracy. Project page: \url{https://ustc3dv.github.io/DualReg/}.
\end{abstract}    
\section{Introduction}
Rigid registration aims to estimate a rigid transformation that aligns a source signal with a target signal. It plays an important role in SLAM, robotics, 3D reconstruction, AR/VR, and 3D data processing. 
Since the signals used for registration are usually partially overlapping and often contain noise and outliers, achieving robust alignment is inherently difficult. Moreover, real-time applications such as SLAM impose strict efficiency requirements. Therefore, developing fast and accurate registration algorithms remains a challenging and critical task. 

Since correspondences are unknown, the Iterative Closest Point (ICP) algorithm~\cite{besl1992method} and its variants~\cite{chen1992object,pavlov2018aa,rusinkiewicz2019symmetric,zhang2021fast} establish correspondences by iteratively searching for closest points in Euclidean space, considering both spatial positions and geometric information of the processed signals. Through alternating optimization of rigid transformations and correspondence establishment on transformed source and target signals, the process gradually converges to an optimal solution. 
These methods enable fine-grained local alignment but depend on the initial spatial positioning between source and target signals. When encountering substantial transformation discrepancies, they frequently converge to suboptimal local minima.

\begin{figure}[t]
  \centering
\includegraphics[width=1\columnwidth]{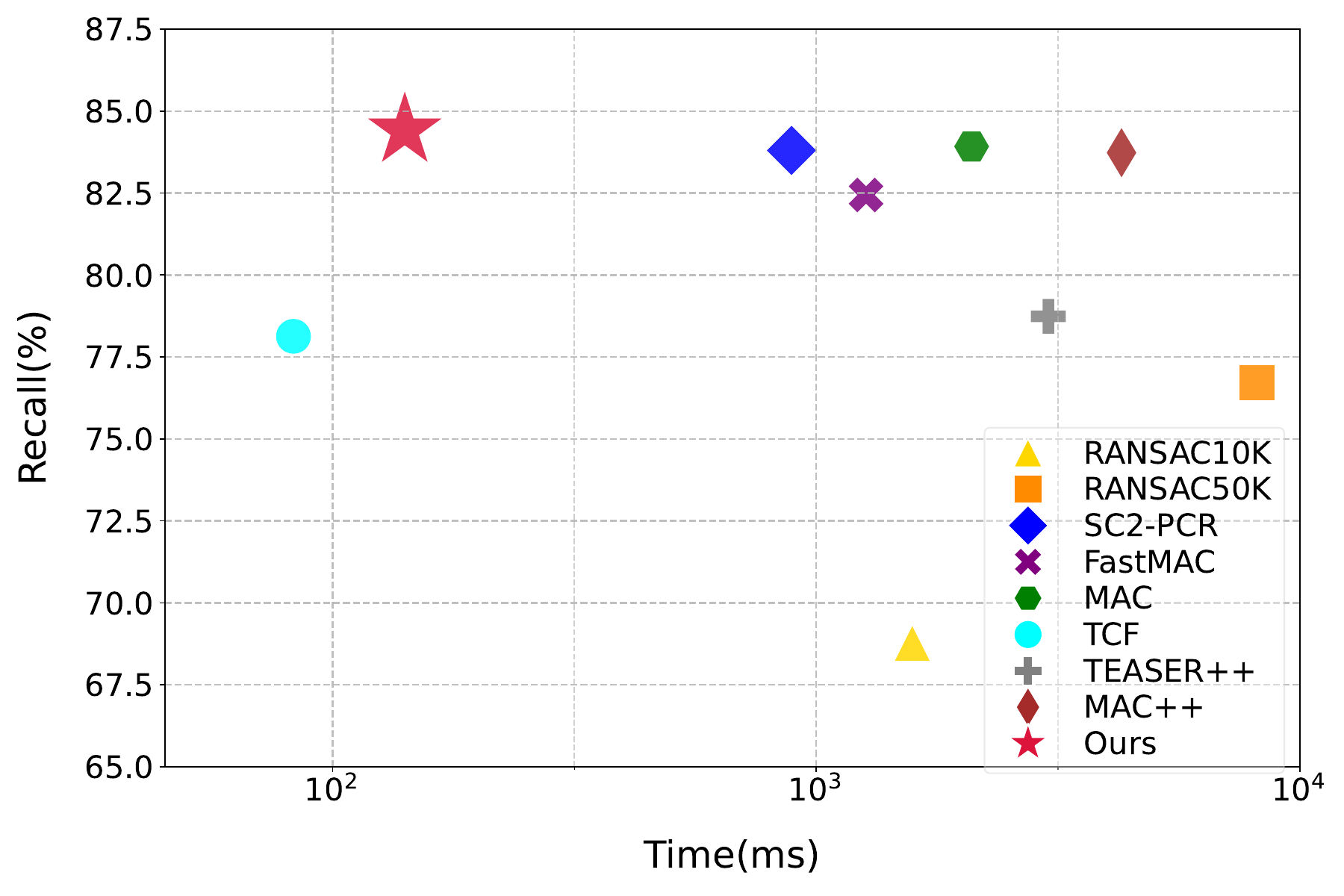} 
  \caption{Registration Recall and Average Runtime (per-registration) on the 3DMatch dataset with FPFH features. Optimal methods require high recall and low runtime. Our method achieves the highest recall with lower time cost. } 
  \label{fig:comparison} 
\end{figure}

In contrast, some methods construct descriptors for input signals and establish feature-based corresponding points globally based on the similarity of descriptors~\cite{rusu2009fast,zhou2016fast,khoury2017learning}.
However, when the data is noisy or its density is uneven, the quality of correspondences is greatly degraded.
With the popularity of deep learning, many methods~\cite{khoury2017learning,deng2018ppfnet,zeng20173dmatch,choy2019fully} design network structures to learn point cloud descriptors. 
Although these methods are dedicated to extracting better feature descriptors to establish better correspondences, the presence of outliers remains inevitable. 
Therefore, numerous outlier rejection methods have been proposed to filter the inaccurate correspondences~\cite{zeng20173dmatch,Parra2018,deng2018ppfnet,gojcic2019perfect,bai2021pointdsc,Lijiayuan2022,yangjiaqi2022}. 
The traditional method is the RANSAC algorithm~\cite{fischler1981random}, which iteratively generates hypotheses
and estimates the model until it finds the subset with the highest inlier rate. 
However, the RANSAC algorithm usually requires a large number of iterations, which may lead to high computation costs. Therefore, some subsequent methods~\cite{torr2002napsac,barath2018graph,quan2020compatibility,yang2021sac,nie2023rlsac} have improved its performance. TCF~\cite{shi2024ransac} proposes one-point RANSAC and two-point RANSAC to greatly reduce the number of iterations, but it reduces alignment quality. 
Recently, MAC~\cite{zhang20233d} constructs a compatibility graph for the initial correspondences and generates hypotheses by searching for the maximal cliques. It greatly improves the registration accuracy, but its computation cost is high. Follow-up works~\cite{zhang2024fastmac,zhang2025mac} have been proposed to improve its speed or accuracy. However, a registration method that is both fast and precise remains essential.

In this paper, we aim to develop an efficient and effective rigid registration algorithm capable of rapidly filtering out inaccurate matches from initial candidate correspondences and estimating precise rigid transformations. We observe that ICP-based methods yield corresponding points in local geometric space, which is fine-grained but depends on the initial transformation. Conversely,  feature-space correspondences obtained through techniques such as~\cite{rusu2009fast,choy2019fully} are transformation-invariant but often fail to achieve precise local alignment. To combine the strengths of both approaches, a feasible solution is to utilize feature-based methods to obtain a coarse rigid transformation, which is then further refined by ICP-based methods to achieve high-precision registration. However, under low-overlap scenarios, ICP-based local registration algorithms often converge to suboptimal solutions due to erroneous nearest point correspondences. While robust norm-based ICP variants~\cite{bouaziz2013sparse,zhang2021fast} can reduce sensitivity to outliers, their fundamental reliance on spatial proximity constraints limits their performance.

To overcome these limitations, we propose a dual-space optimization framework that synergizes correspondences between the global feature and local geometric spaces to achieve high-precision registration. To establish reliable correspondences within the feature space, we design an efficient correspondence filtering strategy that rapidly extracts high-confidence correspondences from raw matches generated using either hand-crafted or learned features. One of our contributions is the design of a computationally lightweight 
1-point RANSAC algorithm for preliminary correspondence filtering, followed by refinement using a 3-point RANSAC algorithm. These high-confidence correspondences enable the construction of geometric proxy point sets with significantly improved overlap with the input point cloud. On these proxy point sets, we establish local geometry-based correspondences via a nearest neighbor search in Euclidean space. We then construct a dual-space optimization framework that synergistically integrates feature- and geometry-based correspondences and design an efficient iterative solver to compute the optimal rigid transformation. Experimental results demonstrate the effectiveness and efficiency of our approach.

In summary, our main contributions include:
\begin{itemize}
\item We design a dual-space optimization framework that synergistically integrates correspondences in feature and local geometric spaces to accurately estimate rigid transformations.
\item We propose an efficient progressive filtering mechanism for feature-based correspondences, which is achieved through fast filtering based on 1-point RANSAC and accuracy refinement using a probability-guided 3-point RANSAC sampler.
\item Based on the filtered feature-based correspondences, we construct a geometric proxy point set and dynamically establish geometry-based correspondences, ultimately achieving accurate rigid registration.
\end{itemize}
\section{Related Work}

In this section, we review the methods relevant to this paper. 

\mypara{Correspondence Prediction}
Extracting accurate correspondences between source data and target data is crucial to achieving global alignment of objects or scenes. Various methods have been proposed to address this challenge. Some approaches select keypoints from point clouds and construct descriptors for them using hand-crafted features, such as FPFH~\cite{rusu2009fast} and SHOT~\cite{salti2014shot}, and establish correspondences between keypoints based on the similarity of their descriptors~\cite{zhou2016fast}. 

Alternatively, many recent methods adopt deep learning techniques to predict more accurate correspondences. LCGF~\cite{khoury2017learning} first extracts hand-crafted features and then maps them to a compact representation using a network. 
3DMatch~\cite{zeng20173dmatch} extracts local 3D patches, converts them into a volumetric representation, and learns 3D ConvNet-based descriptors to establish the correspondences. 
PPFNet~\cite{deng2018ppfnet} adopts multiple PointNets~\cite{qi2017pointnet} and max-pooling aggregation to design a discriminative descriptor. 
FCGF~\cite{choy2019fully} proposes a 3D fully-convolutional network to obtain compact geometric features.
D3Feat~\cite{bai2020d3feat} proposes a joint learning framework for keypoint detection and 3D local feature description.
Predator~\cite{huang2021predator} proposes an overlap-attention block that enables information exchange between latent encodings of two point clouds to deal with partial overlaps, especially in low-overlap scenarios.
GeoTransformer~\cite{qin2023geotransformer} proposes a geometric transformer for point clouds to learn a transformation-invariant nonlocal geometric representation for unsampled superpoint matching.

\mypara{Outlier Removal}
As the initially obtained correspondences may not be entirely accurate, it is necessary to filter them to remove outlier points.
One of the most commonly used methods is based on the Random Sample Consensus (RANSAC) algorithm~\cite{fischler1981random}, which iteratively samples random data points and estimates the model until a satisfactory solution is obtained. 
Numerous improvements have been proposed to enhance RANSAC's performance~\cite{torr2002napsac,barath2018graph,quan2020compatibility,yang2021sac,piedade2023bansac}.
These methods employ weighted sampling strategies to reduce sampling randomness or introduce new compatibility metrics, thereby enhancing efficiency and model estimation accuracy.

Some methods~\cite{lusk2021clipper,fathian2024clipper+,zhang20233d,zhang2024fastmac,zhang2025mac,yan2025turboreg} obtain high-confidence inliers by introducing a graph and finding maximal cliques.
MAC~\cite{zhang20233d} utilizes the second-order spatial compatibility measure proposed by SC$^2$-PCR~\cite{chen2022sc2} to construct a compatibility graph and proposes a new strategy to search for maximal cliques efficiently. Subsequently, FastMAC~\cite{zhang2024fastmac} accelerates MAC by filtering the graph. 
MAC++~\cite{zhang2025mac} improves upon MAC by constructing a maximal clique pool for each node and evaluating hypotheses in a global-to-local manner to boost registration recall in extreme cases.
CLIPPER+~\cite{fathian2024clipper+} proposes the Degeneracy-Ordered Greedy algorithm to improve the speed and accuracy of solving the maximal clique problem.
TurboReg~\cite{yan2025turboreg} introduces TurboClique as a 3-clique within a highly-constrained compatibility graph to improve the efficiency of the maximal clique search.

Moreover, learning-based methods have also been proposed to filter out inaccurate correspondences. 
3DRegNet~\cite{pais20203dregnet} 
proposes a classification block using ResNet to classify inliers and outliers. 
DGR~\cite{choy2020deep} introduces a convolutional network to estimate the veracity of each candidate correspondence. 
PointDSC~\cite{bai2021pointdsc} proposes a spatial-consistency guided nonlocal module and a differentiable neural spectral matching module to learn reliable correspondences. 
VBReg~\cite{jiang2023vbreg} develops an outlier rejection framework based on variational non-local networks and aggregates discriminative geometric context information through Bayesian-driven long-range dependencies to distinguish inliers and outliers. 
HyperGCT~\cite{zhang2025hypergct} constructs a hypergraph and utilizes Hyper-GNN to learn high-order correlations, resulting in geometric constraints that are robust to noise.
HeMoRa~\cite{yan2025hemora} designs a multi-order
reward aggregator loss to learn the sampling probabilities of correspondences in an unsupervised manner.

\mypara{Robust Estimation of Transformation}
For given correspondences, classical methods typically minimize the $\ell_2$-norm of the alignment error to estimate the transformation~\cite{besl1992method}. However, outliers or incorrect correspondences can severely degrade the accuracy of the results.
Therefore, a large number of methods adopt robust objective functions to enhance the accuracy of the registration.
Sparse-ICP~\cite{bouaziz2013sparse} uses $\ell_p$-norm ($0<p<1$); FGR~\cite{zhou2016fast} uses Geman-McClure; 
TEASER++~\cite{yang2021teaser} applies truncated least squares with graph-theoretic decomposition. 
FRICP~\cite{zhang2021fast} employs the Welsch function~\cite{holland1977robust} with Anderson acceleration~\cite{walker2011anderson} for rigid registration. 
The Welsch function also benefits non-rigid registration via a quasi-Newton solver~\cite{Yao_2020_CVPR} and an accelerated MM algorithm~\cite{yao2023fast}, ensuring robustness and smoothness. 
By leveraging random line intersections, a robust loss~\cite{deng2021robust} is proposed to enhance the accuracy of correspondences.
TCF~\cite{shi2024ransac} uses a Cauchy objective, while Jiang et al.~\cite{jiang2024l0minimization} formulate inlier identification as $\ell_0$ minimization. 
\section{Problem Statement and Method Overview}

\begin{figure*}[!htb]
  \centering
\includegraphics[width=0.8\textwidth]{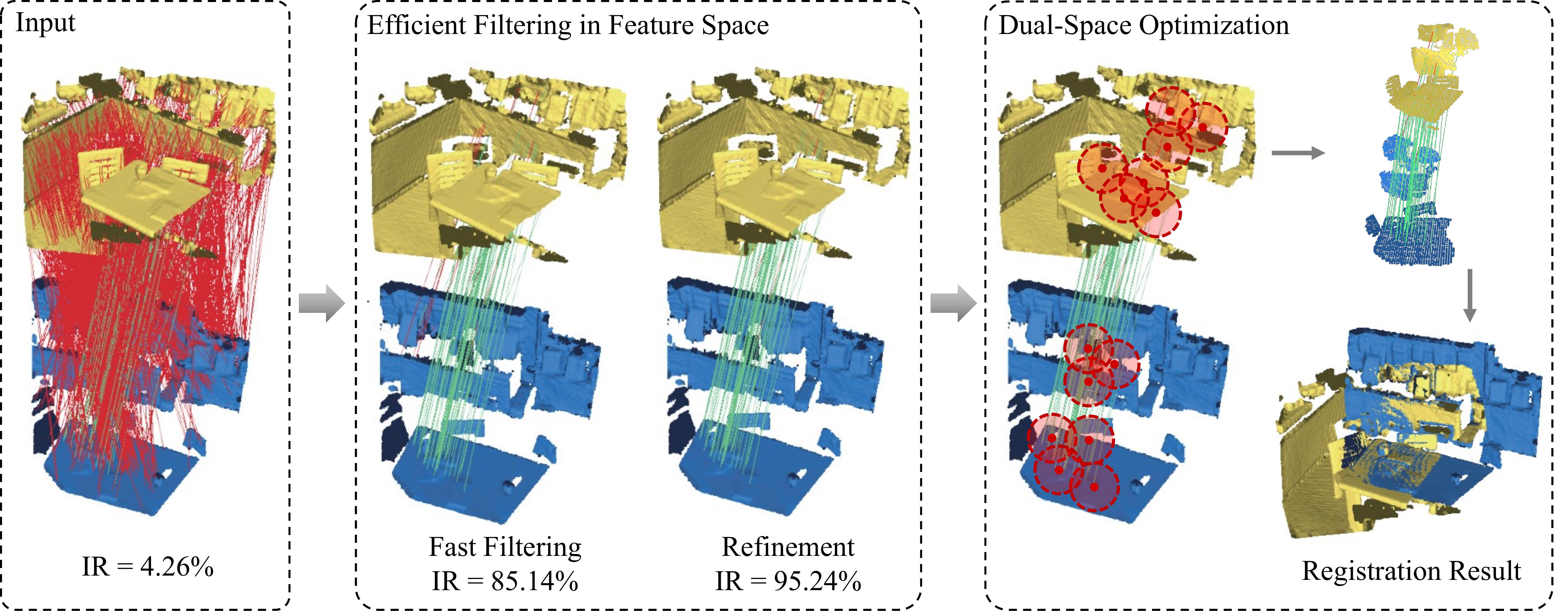} 
  \caption{Pipeline of the proposed method. We propose DualReg, a dual-space paradigm for robust rigid registration. We first propose an efficient filtering algorithm for feature-based correspondences including 1-point RANSAC-based fast filtering and 3-point RANSAC-based refinement. Then, according to the filtered feature-based correspondences, we construct geometric proxies and propose a dual-space optimization framework to jointly estimate the rigid transformation.  
  We mark the inliers and outliers with green and red lines, respectively, and mark the ground truth inlier rate (IR) at each stage.}
  \label{fig:pipeine} 
\end{figure*}

Given source data, represented as a point set \( \mathcal{V} = \{ \mathbf{v}_1,...,\mathbf{v}_{|\mathcal{V}|}\in\mathbb{R}^3\} \) with corresponding normals $\mathcal{N}_s=\{\mathbf{n}^s_1,...,\mathbf{n}^s_{|\mathcal{V}|}\}$ and target data, represented as a point set $\mathcal{U} = \{ \mathbf{u}_1,...,\mathbf{u}_{|\mathcal{U}|}\in\mathbb{R}^3\}$ with corresponding normals $\mathcal{N}_t=\{\mathbf{n}^t_1,...,\mathbf{n}^t_{|\mathcal{U}|}\}$, the goal of rigid registration is to compute a rigid transformation $\mathbf{T}^*=[\mathbf{R}^*, \mathbf{t}^*]$ so that 
\[
(\mathbf{R}^*, \mathbf{t^*}) = \mathop{\arg\min}_{(\mathbf{R},\mathbf{t})} \sum_{(\mathbf{v}_j, \mathbf{u}_j)\in\mathcal{C}}\mathcal{D}(\mathbf{R}\mathbf{v}_j + \mathbf{t} - \mathbf{u}_j),
\]
where $\mathbf{R}\in\texttt{SO(3)}$ is a rotation matrix and $\mathbf{t}\in\mathbb{R}^3$ is a translation vector. 
$\mathcal{D}(\cdot)$ is a metric to measure the alignment quality between the corresponding points, which typically employs the $\ell_2$-norm or other robust functions~\cite{bouaziz2013sparse,zhou2016fast,zhang2021fast}.
The correspondence set $\mathcal{C}$ serves as a fundamental component in point cloud registration, playing a pivotal role in estimating optimal rigid transformations. 
We observe that correspondences are typically derived from two distinct spaces:
\begin{itemize}
\item \textbf{Feature Space:} correspondences are constructed by extracting transformation-invariant global features, such as FPFH~\cite{rusu2009fast} and FCGF~\cite{choy2019fully}; 
\item \textbf{Local Geometric Space:} correspondences are formed through closest-point search based on spatial positions and geometric attributes of point clouds, as implemented in ICP~\cite{besl1992method} and its variants~\cite{bouaziz2013sparse,pavlov2018aa,zhang2021fast}.
\end{itemize}
Correspondences derived from the feature space effectively handle significant rotational discrepancies between source and target point clouds, but they lack precision in alignment. While correspondences originating from the local geometric space enable refined registration, they rely on the initial transformation. To leverage these complementary advantages, we propose a novel dual-space paradigm that simultaneously incorporates feature-space correspondences and local geometric-space correspondences,  
achieving efficient and accurate point cloud registration.

Specifically, as illustrated in Fig.~\ref{fig:pipeine}, we first obtain raw correspondences through feature-based methods~\cite{rusu2009fast,choy2019fully}. To remove outliers and obtain correspondences with a higher inlier ratio, we propose an efficient filtering algorithm comprising a fast filtering stage followed by refinement, as detailed in Sec.~\ref{sec:filter-feature-corr}. Subsequently, we construct geometric proxies and formulate a novel dual-space optimization framework in Sec.~\ref{sec:optimization-framework} that simultaneously incorporates both the filtered feature-based correspondences and progressively refined geometry-based correspondences across iterations, ultimately achieving high-precision rigid registration.

\section{Efficient Filtering for Feature-based Correspondences}
\label{sec:filter-feature-corr}
The raw correspondences $\mathcal{C}_0=\{ \mathbf{c}_j = ( \mathbf{v}_j, \mathbf{u}_j )| \mathbf{v}_j\in\mathcal{V}, \mathbf{u}_j\in\mathcal{U}\}$ extracted by feature-based methods often contain many outliers.
The three-point RANSAC algorithm is widely used to obtain consensus subsets~\cite{shi2024ransac, dai2022multisource, chung2024centralized, salehi2022improved, zhang2024improved}. However, too many samples will greatly increase the number of iterations of the RANSAC algorithm. We notice that when we only take one point as a sample, the total number of samples is greatly reduced (from $\mathcal{O}(n^3)$ to $\mathcal{O}(n)$), thereby significantly speeding up the solution. 
Therefore, we propose an efficient 1-point RANSAC algorithm to prune a significant number of erroneous correspondences. This initial filtering step drastically reduces the problem size, enabling a subsequent 3-point RANSAC to achieve higher accuracy with far fewer iterations.
 
\subsection{Determining 1-Point-Based Consensus Set} 
\label{sec:consensus-defination}
For a randomly sampled correspondence $\mathbf{c}_j = (\mathbf{v}_j, \mathbf{u}_j)$, we define a consistency set based on the point positions and normals: 
\begin{equation}
\label{eq:init-one-point-consensus}
\mathcal{I}_{\text{init}}(\mathbf{c}_j) = \{\mathbf{c}_i\in\mathcal{C}_0~|~D_L(\mathbf{c}_i,\mathbf{c}_j)<\tau~\text{and}~D_N(\mathbf{c}_i,\mathbf{c}_j)<\nu\}
\end{equation}
where
\begin{equation}
\label{eq:length-consistent}
D_L(\mathbf{c}_i,\mathbf{c}_j) = \left|\|\mathbf{v}_i-\mathbf{v}_j\|_2 - \|\mathbf{u}_i - \mathbf{u}_j\|_2\right|
\end{equation}
denotes the length differences between the correspondences. 
$\tau$ and $\nu$ are the density and scale-related thresholds for point clouds.
\begin{equation}
\label{eq:normal-consistent}
D_N(\mathbf{c}_i,\mathbf{c}_j) = \max(|d_{ij}^s - d_{ij}^t|, |d_{ji}^s - d_{ji}^t|) 
\end{equation}
is the tangential distance,  
where 
\begin{equation}
\begin{aligned}
&d_{ij}^s = |(\mathbf{v}_i-\mathbf{v}_j)\cdot \mathbf{n}_{i}^s|,  ~~d_{ji}^s = |(\mathbf{v}_j-\mathbf{v}_i)\cdot \mathbf{n}_{j}^s|,\\
&d_{ij}^t = |(\mathbf{u}_i-\mathbf{u}_j)\cdot \mathbf{n}_{i}^t|,  ~~d_{ji}^t = |(\mathbf{u}_j-\mathbf{u}_i)\cdot \mathbf{n}_{j}^t|
\end{aligned}
\end{equation}
are the point-to-plane distances between these points. 
Since normal estimation for point clouds may contain orientation errors, this approach is more robust than directly computing angular differences between normals.

To further obtain a set of correspondences with higher consistency, we select a subset $\mathcal{I}(\mathbf{c}_j)$ of pairwise consensus from $\mathcal{I}_{\text{init}}(\mathbf{c}_j)$, that is 
\begin{equation}
\label{eq:one-point-consensus-set}
\mathcal{I}(\mathbf{c}_j) = \{\mathbf{c}_i\in\mathcal{I}_{\text{init}}(\mathbf{c}_j)~|~\forall \mathbf{c}_k, D_L(\mathbf{c}_i, \mathbf{c}_k) < \tau\}. 
\end{equation}

\begin{figure}[!htb]
  \centering
\includegraphics[width=1\columnwidth]{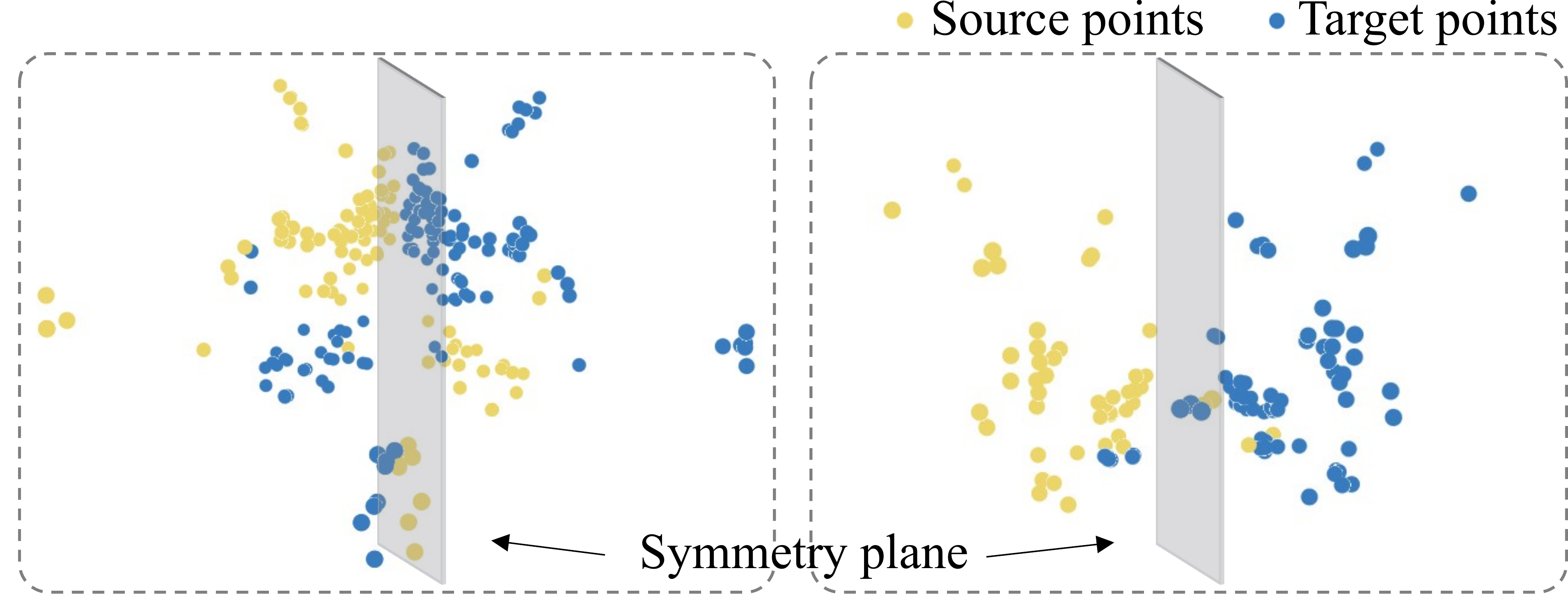} 
  \caption{Correspondences meet symmetry. 
  } 
  \label{fig:symm-corres} 
\end{figure}
\mypara{Disambiguating symmetry}
We note that the above set $\mathcal{I}(\mathbf{c}_j)$ considers only pairwise relationships between corresponding points, neglecting global structure. 
When the source points $\{\mathbf{v}_i\}_{\mathbf{c}_i\in\mathcal{I}(\mathbf{c}_j)}$ and target points $\{\mathbf{u}_i\}_{\mathbf{c}_i\in\mathcal{I}(\mathbf{c}_j)}$ are symmetric about a plane (see Fig.~\ref{fig:symm-corres}), the rigid transformation in 3D space cannot align them.
Therefore, we have to filter out these cases. We compute the optimal orthogonal transformation that aligns the corresponding points. If this transformation is a reflection rather than a rotation, then the correspondence exhibits symmetry. The specific details can be found in the \textit{Supplementary Material}.

\subsection{Iteratively Searching Optimal Consensus Set}
\label{sec:Searching-Optimal-Consensus-Set}
\begin{figure}[!htb]
  \centering
\includegraphics[width=\columnwidth]{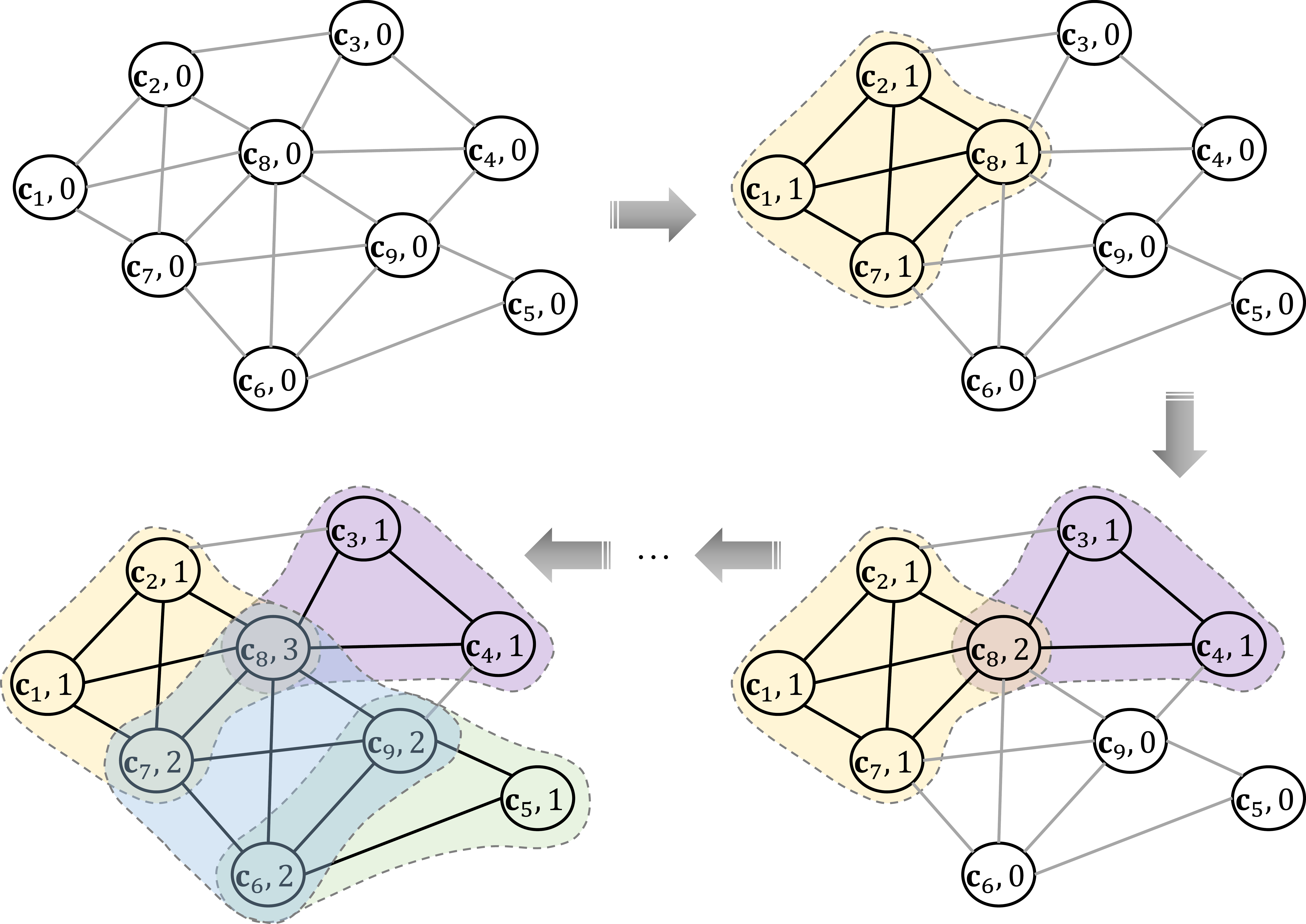} 
  \caption{Iterative process of one-point RANSAC. Each circle in the figure represents a correspondence, labeled in the form of ``$\mathbf{c}_j,s_{\mathbf{c}_j}^{(k)}$", 
  where ``$\mathbf{c}_j$" denotes the $j$-th correspondence and ``$s_{\mathbf{c}_j}^{(k)}$" represents the confidence score of $\mathbf{c}_j$ at $k$-iteration. 
  During the iteration, a correspondence is randomly selected each time to construct its consensus set: if there is a line between two correspondences, it indicates that they satisfy both length consistency and normal consistency; correspondences covered with the same color belong to the same consensus set.} 
  \label{fig:one-point-iterate} 
\end{figure}

To estimate the accuracy of the corresponding points,
we define a confidence score $s_{\mathbf{c}_j}$ for each correspondence $\mathbf{c}_j\in\mathcal{C}_0$, and propose a one-point RANSAC-based confidence estimation algorithm. The score is initialized to 0 and dynamically updated by counting the frequency of $\mathbf{c}_j$ appearing in high-quality consistent sets. Higher frequency indicates stronger robustness to geometric constraints, with the final $s_{\mathbf{c}_j}$ quantifying correspondence reliability.
Furthermore, we use $N_{\text{I}}^{(k)}$ to denote the number of estimated inliers in $\mathcal{C}_0$ at the $k$-th iteration. 
Initially, we set $s_{\mathbf{c}_j}^{(0)}=0$ for all $\mathbf{c}_j$ and $N_{{\text{I}}}^{(0)}=1$. 
At the $k$-th iteration, we first randomly sample a correspondence $\mathbf{c}_j^{(k)}\in\mathcal{C}_0$, then compute the consensus set $\mathcal{I}(\mathbf{c}_j^{(k)})$ for $\mathbf{c}_j^{(k)}$. 
Since we assume that a larger consensus set is more likely to be an inlier set, we consider $\mathcal{I}(\mathbf{c}_j^{(k)})$ as a reliable set when 
\[
|\mathcal{I}(\mathbf{c}_j^{(k)})| > \alpha \cdot N_{\text{I}}^{(k-1)},
\]
where $\alpha$ is a user-defined parameter and $|\cdot|$ denotes the number of elements in the set. 
In this case, we will update the confidence score $s_{\mathbf{c}_i}^{(k)} = s_{\mathbf{c}_i}^{(k-1)}+1$ for each correspondence $\mathbf{c}_i\in\mathcal{I}(\mathbf{c}_j^{(k)})$. We approximate the number of inliers by the number of maximum consensus sets in the current iteration, that is 
\[
N_{\text{I}}^{(k)}=\max(N_{\text{I}}^{(k-1)},~ |\mathcal{I}(\mathbf{c}_j^{(k)})|).
\]
Fig.~\ref{fig:one-point-iterate} illustrates the iterative procedure.

\mypara{Termination Criteria} 
Refer to~\cite{shi2024ransac}, we dynamically adjust the total number of iterations $K_{\text{I}}^{(k)}$ based on the current estimated number of inliers: 
\begin{equation}
\label{eq:one-point-ransac-termination}
K_{\text{I}}^{(k)} = \left\lceil\frac{\log (1-\delta)}{\log(1-N_{\text{I}}^{(k)}/|\mathcal{C}_0|)}\right\rceil, 
\end{equation}
where $\lceil\cdot\rceil$ is a ceiling function and $\delta=0.99$ by default.

\mypara{Selecting the Best Consensus Set}
After terminating the iteration, we denote the final score of each correspondence $\mathbf{c}_j$ as $s_{\mathbf{c}_j}^*$, and save the consensus set $\mathcal{I}(\mathbf{c}^{(k)}_j)$ during the iterations. Then we compute the score sum of each correspondence in $\mathcal{I}(\mathbf{c}_j^{(k)})$ and obtain the consensus set with the highest total score:
\begin{equation}
\label{eq:best-C-coarse}
\mathcal{I}(\mathbf{c}_*) = \argmax_{\mathcal{I}(\mathbf{c}_j^{(k)})} \sum_{\mathbf{c}_i\in\mathcal{I}(\mathbf{c}^{(k)}_j)} s_{\mathbf{c}_i}^*. 
\end{equation}
It will enter the following process as the best consensus set after coarse filtering. We also denote $\mathcal{I}(\mathbf{c}_*)$ as $\mathcal{C}_{\text{I}}$.

\vspace{0.2em}
\noindent\textbf{\textit{Remark.}}
Although TCF~\cite{shi2024ransac} introduce one-point RANSAC for acceleration, our method fundamentally differs in three aspects: (1) the definition of one-point-based consensus set;
(2) the optimal subset selection criteria, and (3) the symmetry checking module.

\subsection{Refinement with Weighted 3-Point Sampling}
\label{sec:refinement}
Since outliers may also satisfy the length and tangential distance constraints described in Eq.~\eqref{eq:one-point-consensus-set}, we need to perform more detailed filtering on $\mathcal{C}_{\text{I}}$.
Inspired by~\cite{piedade2023bansac}, we propose a refinement algorithm based on a probability-based weighted sampling strategy.
We assign an inlier probability to each correspondence, dynamically update the probability according to a dynamic Bayesian network proposed by~\cite{piedade2023bansac}, and perform the probability-based weighted sampling. Compared with the classic RANSAC algorithm, it reduces many unnecessary iterations, thereby reducing the running time while maintaining high accuracy. Specific details are presented in the \textit{Supplementary Material}.
The best transformation $(\mathbf{R}_{\text{II}}^*,\mathbf{t}_{\text{II}}^*)$ and the inlier set $\mathcal{C}_{\text{II}}$ will be obtained to perform robust transformation estimation.

\section{Dual-Space Optimization Framework}
\label{sec:optimization-framework}
The filtered correspondence set $\mathcal{C}_{\text{II}}$ has a higher inlier rate than the initial $\mathcal{C}_0$, but there may be problems such as the corresponding points cannot be accurately aligned under the transformation, and $\mathcal{C}_{\text{II}}$ still has outliers.
To obtain an accurate rigid transformation to align the source point cloud and target point cloud, we propose an additional robust geometry-aware estimation algorithm.  
\subsection{Construction of Geometric Proxies}
\label{sec:Geometric-Proxies}
We regard the corresponding points in $\mathcal{C}_{\text{II}}$ as anchor points and extract the local neighbors on the point clouds as proxy points on the potential surfaces.  Specifically, for $\mathbf{c}_j=(\mathbf{v}_j,\mathbf{u}_j)\in\mathcal{C}_{\text{II}}$, we construct the proxy point sets for $\mathbf{c}_j$ as:  
\begin{equation}
\begin{aligned}
&\mathcal{P}^s_{\mathbf{c}_j} = \{\mathbf{v}_i\in\mathcal{V}~|~\|\mathbf{v}_i - \mathbf{v}_j\|_2<\beta\} \\
&\mathcal{P}^t_{\mathbf{c}_j} = \{\mathbf{u}_i\in\mathcal{U}~|~\|\mathbf{u}_i - \mathbf{u}_j\|_2<\beta\}.
\end{aligned}
\end{equation}
Through the neighbor aggregation process, we can obtain 
\begin{equation}
\mathcal{P}^s = \{\mathcal{P}_{\mathbf{c}_j}^s\}_{\mathbf{c}_j\in\mathcal{C}_{\text{II}}}, \quad 
\mathcal{P}^t = \{\mathcal{P}_{\mathbf{c}_j}^t\}_{\mathbf{c}_j\in\mathcal{C}_{\text{II}}}. 
\end{equation}
Compared with the input point clouds $\mathcal{V}$ and $\mathcal{U}$, $\mathcal{P}^s$ and $\mathcal{P}^t$ have a significantly higher overlap rate (refer to Fig.~\ref{fig:pipeine}).
We utilize the spatial geometric information of $\mathcal{P}^s$ and $\mathcal{P}^t$ in combination with $\mathcal{C}_{\text{II}}$ to estimate the rigid transformation, avoiding errors caused by fewer and inaccurate correspondences. 

\subsection{ Optimization Formulation}
\label{sec:Optimization-Formulation}
We propose the following objective function:
\begin{equation}
\begin{aligned}
\label{eq:RT-post-opt}
E(\mathbf{R}, \mathbf{t}) =&
~\frac{\lambda}{|\mathcal{C}_{\text{II}}|} \cdot  \sum_{\mathbf{c}_j \in \mathcal{C}_{\text{II}}} w_j\|\mathbf{R} \mathbf{v}_j + \mathbf{t} - \mathbf{u}_j\|^2  \\
&+ \frac{1}{|\mathcal{P}^s|}\cdot\sum_{\widetilde{\mathbf{v}}_i \in \mathcal{P}^s} \widetilde{w}_i  \|\mathbf{R} \widetilde{\mathbf{v}}_i + \mathbf{t} - \widetilde{\mathbf{u}}_{\rho_i}\|^2,
\end{aligned}
\end{equation}
where $\widetilde{\mathbf{u}}_{\rho_i}\in\mathcal{P}^t$ is the corresponding point for $\widetilde{\mathbf{v}}_i$. $\lambda$ is a parameter to balance the two terms. 
$w_j$ and $\widetilde{w}_i$ are the robust weights to avoid the effect of inaccurate corresponding points, and can be computed by the following function: 
\begin{equation}
\label{eq:robust_weight}
\mathcal{W}(\mathbf{R}, \mathbf{t}, \mathbf{v},\mathbf{u}) = \exp\left(-\frac{\|\mathbf{R}\mathbf{v} + \mathbf{t} - \mathbf{u}\|^2}{2\sigma^2}\right),
\end{equation}
where $\sigma$ is a parameter. 
It uses the Gaussian function to reduce the impact of large residual values, which is similar to~\cite{yao2025spare}. 
According to the classic 3-sigma rule, correspondences with distances exceeding 3$\sigma$ will be heavily down-weighted. 
Leveraging the estimated probability $p_{\mathbf{c}_j}$ defined in Sec.~\ref{sec:refinement}, we select a high-confidence subset $\widehat{\mathcal{C}}_{\text{II}}\subset \mathcal{C}_{\text{II}}$  where $|\widehat{\mathcal{C}}_{\text{II}}|/|\mathcal{C}_{\text{II}}|=0.4$ by default. The parameter $\sigma$ is then computed as:
$\sigma=\frac{1}{3} \max_{\mathbf{c}_j \in \widehat{\mathcal{C}}_{\text{II}}} \|\mathbf{R}_{\text{II}}^* \mathbf{v}_j +\mathbf{t}_{\text{II}}^* - \mathbf{u}_j\|$.

\subsection{Numerical Solver}
Since it is difficult to directly solve $\{\rho_i\}$ considering the spatial geometry, we adopt the idea of ICP~\cite{besl1992method} and iteratively construct $\{\rho_i\}$ and solve $(\mathbf{R},\mathbf{t})$. Initially, we set $\mathbf{R}^{(0)}=\mathbf{R}_{\text{II}}^*$, $\mathbf{t}^{(0)} = \mathbf{t}_{\text{II}}^*$ and then perform the following steps at the $k$-th iteration:
\begin{itemize}
\item We fix transformation $(\mathbf{R}^{(k-1)}, \mathbf{t}^{(k-1)})$ and update the indices of the corresponding points $\{\rho_i\}$ by performing a closest-point search:
\begin{equation}
\widetilde{\mathbf{u}}_{\rho_i^{(k)}} = \argmin_{\widetilde{\mathbf{u}}\in\mathcal{P}^t} \|\mathbf{R}^{(k-1)}\widetilde{\mathbf{v}}_i + \mathbf{t}^{(k-1)}-\widetilde{\mathbf{u}}\|_2.
\end{equation}
\item We fix the correspondences $\{\widetilde{\mathbf{u}}_{\rho_i^{(k)}}\}$ and transformation $(\mathbf{R}^{(k-1)}, \mathbf{t}^{(k-1)})$ to compute the robust weights $\{w_j^{(k)}\}, \{\widetilde{w}_i^{(k)}\}$ by Eq.~\eqref{eq:robust_weight}.
\item  We fix the correspondences $\{\widetilde{\mathbf{u}}_{\rho_i^{(k)}}\}$ and robust weight $\{w_j^{(k)}\}, \{\widetilde{w}_i^{(k)}\}$ to solve the rigid transform $(\mathbf{R}^{(k)}, \mathbf{t}^{(k)})$ by optimizing the objective function in Eq.~\eqref{eq:RT-post-opt}. 
It can be solved in closed form via SVD~\cite{RigidSVD}. 
\end{itemize}
We terminate the iteration when $\|\mathbf{T}^{(k)}-\mathbf{T}^{(k-1)}\|_F < \epsilon$ or $k>K_{\text{III}}$. Here $\mathbf{T}^{(k)}=[\mathbf{R}^{(k)}, \mathbf{t}^{(k)}]$ and we set $\epsilon=0.001, K_{\text{III}}=200$ by default.  The final rigid transformation $(\mathbf{R}^*, \mathbf{t}^*)$ can be obtained. 
\section{Results}
In this section, we first introduce our experimental setup, then present a comparison with the state-of-the-art methods, and finally test the effectiveness of each part of our method in outlier removal and transformation estimation. Further effectiveness analyses and visualizations can be found in the \textit{supplementary material} and \textit{demo video}.

\mypara{Experimental Setup}
To validate the performance of our method, we followed the setup in~\cite{zhang20233d,yan2025turboreg}, and tested on two indoor datasets, 3DMatch~\cite{zeng20173dmatch} and  3DLoMatch~\cite{huang20213dlomatch}, and one outdoor dataset, KITTI~\cite{geiger2012KITTI}.
The 3DLoMatch dataset serves as a complementary benchmark to 3DMatch, containing point cloud pairs with low overlap ratios (10\%–30\%), thereby posing a more challenging registration scenario.
We compared the proposed method with optimization-based methods: RANSAC10K~\cite{fischler1981random}, RANSAC50K~\cite{fischler1981random}, TEASER++~\cite{yang2021teaser}, SC2-PCR~\cite{chen2022sc2}, MAC~\cite{zhang20233d}, FastMAC~\cite{zhang2024fastmac}, TCF~\cite{shi2024ransac}, MAC++~\cite{zhang2025mac} and TurboReg~\cite{yan2025turboreg}, as well as two learning-based methods: PointDSC~\cite{bai2021pointdsc} and VBReg~\cite{jiang2023vbreg}. 
Our method was implemented in C++ with Point Cloud Library (PCL) and  Eigen library on Ubuntu 20.04. 
All learning-based methods and TurboReg were executed on a single NVIDIA RTX 3090 GPU.
For a fair comparison, all other optimization-based methods were run with GPU usage disabled on 
a 6-core/12-thread Intel\textregistered{} Core\texttrademark{} i7-8700 CPU@3.20GHz.  
We evaluated the performance on all datasets using registration recall (RR), root
mean square error (RMSE), rotation error (RE), and translation error (TE), referring to~\cite{choy2020deep,zhang2021fast,zhang20233d,yan2025turboreg}. 
In the tables, we used bold and underline to mark the best and second-best values, respectively. In particular, GPU runtime was excluded from our comparison to ensure fairness.
More details are shown on the \textit{Supplementary Material}.

\begin{table*}[t]
\centering
\setlength{\tabcolsep}{1.8pt} 
\caption{Performance comparison on the 3DMatch dataset, 3DLoMatch dataset and KITTI dataset.}
\label{tab:compare_all}
\footnotesize
\begin{tabular}{@{}cl|ccccc|ccccc@{}}
\toprule
 &\multirow{2}{*}{\textbf{Method}} & 
\multicolumn{5}{c|}{\textbf{FPFH}} & 
\multicolumn{5}{c}{\textbf{FCGF}} \\
\cmidrule(lr){3-7} \cmidrule(lr){8-12}
& & 
$\mathbf{RR}^{\uparrow} (\%) $ & 
$\mathbf{RMSE}^{\downarrow} (\si{cm})$ & 
$\mathbf{RE}^{\downarrow} (\si{\degree}) $ & 
$\mathbf{TE}^{\downarrow} (\si{cm}) $ & 
$\mathbf{Time}^{\downarrow} (\si{s}) $ & 
$\mathbf{RR}^{\uparrow} (\%) $ & 
$\mathbf{RMSE}^{\downarrow} (\si{cm})$ & 
$\mathbf{RE}^{\downarrow} (\si{\degree}) $ & 
$\mathbf{TE}^{\downarrow} (\si{cm}) $ & 
$\mathbf{Time}^{\downarrow} (\si{s}) $ \\
\midrule
\multirow{14}{*}{\rotatebox{90}{3DMatch}} 
& PointDSC~\cite{bai2021pointdsc} & 78.00 & 4.82 & \underline{2.08} & \underline{6.48} & 0.04$^{*}$ & 92.85 & 4.75 & 2.04 & 6.50 & 0.05$^{*}$ \\
& VBReg~\cite{jiang2023vbreg} & 81.39 & \underline{4.77} & 2.12 & 6.64 & 0.18$^{*}$ & 93.47 & \underline{4.73} & 2.04 & 6.49 & 0.19$^{*}$ \\
\cmidrule(l){2-12}
& RANSAC10K~\cite{fischler1981random} & 68.76 & 8.14 & 3.86 & 10.73 & 1.58 & 91.56 & 5.88 & 2.73 & 8.31 & 1.55 \\
& RANSAC50K~\cite{fischler1981random} & 76.71 & 7.02 & 3.34 & 9.46 & 8.16 & 92.05 & 5.59 & 2.52 & 7.84 & 7.99 \\
& TEASER++~\cite{yang2021teaser} & 78.74 & 5.56 & 2.52 & 7.55 & 3.02 & 85.64 & 5.64 & 2.75 & 9.26 & 8.53 \\
& SC2-PCR~\cite{chen2022sc2} & 83.80 & 4.88 & 2.12 & 6.60 & 0.89 & 92.98 & 4.81 & 2.05 & 6.47 & 0.89 \\
& FastMAC~\cite{zhang2024fastmac} & 82.44 & 4.94 & 2.11 & 6.68 & 1.27 & 92.61 & 4.74 & \underline{2.01} & \underline{6.43} & 1.26 \\
& MAC~\cite{zhang20233d} & 83.92 & 4.94 & 2.11 & 6.79 & 2.10 & \textbf{93.72} & 4.78 & 2.03 & 6.54 & 1.82 \\
& TCF~\cite{shi2024ransac} & 78.13 & 5.32 & 2.37 & 7.54 & \textbf{0.08} & 88.79 & 5.04 & 2.19 & 7.24 & \underline{0.80} \\
& MAC++~\cite{zhang2025mac} & 83.73 & 4.78 & 2.11 & 6.78 & 4.28
 & 93.22 & 4.91 & 2.10 & 6.83 & 5.19
\\
& TurboReg~\cite{yan2025turboreg} & \underline{84.10} & 4.99 & 2.19 & 6.81 & 0.03$^*$ & 92.98 & 4.75 & 2.05 & 6.53 & 0.03$^*$\\
& Ours & \textbf{84.41} & \textbf{4.55} & \textbf{1.75} & \textbf{6.19} & \underline{0.14} & \underline{93.65} & \textbf{4.44} & \textbf{1.65} & \textbf{6.13} & \textbf{0.45} \\
\midrule
\multirow{14}{*}{\rotatebox{90}{3DLoMatch}} 
& PointDSC~\cite{bai2021pointdsc} & 28.86 & \underline{8.68} & \underline{3.92} & 10.27 & 0.05$^{*}$ & 56.65 & 8.72 & 3.79 & 10.42 & 0.05$^{*}$ \\
& VBReg~\cite{jiang2023vbreg} & 37.00 & 8.89 & 4.04 & \underline{10.20} & 0.19$^{*}$ & 59.74 & 8.85 & 3.80 & 10.46 & 0.18$^{*}$ \\
\cmidrule(l){2-12}
& RANSAC10K~\cite{fischler1981random} & 18.75 & 13.67 & 6.53 & 14.35 & 1.22 & 49.02 & 10.61 & 4.74 & 12.36 & 1.39 \\
& RANSAC50K~\cite{fischler1981random} & 26.56 & 11.73 & 5.55 & 12.92 & 6.74 & 54.80 & 10.33 & 4.51 & 12.03 & 7.32 \\
& TEASER++~\cite{yang2021teaser} & 28.30 & 10.16 & 4.75 & 11.64 & 1.80 & 40.43 & 10.19 & 4.82 & 13.68 & 2.47 \\
& SC2-PCR~\cite{chen2022sc2} & 39.53 & 9.22 & 4.11 & 10.39 & 0.85 & 58.06 & 8.70 & \underline{3.75} & 10.44 & 0.86 \\
& FastMAC~\cite{zhang2024fastmac} & 38.18 & 9.25 & 4.12 & 10.45 & 1.03 & 57.72 & 8.96 & 3.80 & 10.62 & 0.70 \\
& MAC~\cite{zhang20233d} & 41.10 & 9.18 & 4.03 & 10.46 & 1.85 & 60.13 & 8.87 & \underline{3.75} & 10.59 & 1.97 \\
& TCF~\cite{shi2024ransac} & 32.57 & 9.66 & 4.26 & 11.38 & \textbf{0.06} & 48.91 & 9.53 & 4.16 & 11.25 & \textbf{0.12} \\
& MAC++~\cite{zhang2025mac} & \textbf{44.58} & 9.54 & 4.28 & 11.38 & 3.29
 & \textbf{61.09} & 9.29 & 3.99 & 11.39 & 3.72\\
& TurboReg~\cite{yan2025turboreg} & 39.19 & 8.95 & 4.03 & 10.36 & 0.02$^*$ & 58.34 & \underline{8.65} & 3.76 & \underline{10.35} & 0.03$^*$\\ 
& Ours & \underline{41.66} & \textbf{7.98} & \textbf{3.08} & \textbf{9.36} & \underline{0.11} & \underline{60.86} & \textbf{7.81} & \textbf{3.06} & \textbf{9.61} & \underline{0.16} \\
\midrule 
\multirow{14}{*}{\rotatebox{90}{KITTI}} 
& PointDSC~\cite{bai2021pointdsc} & 97.12 & 17.36 & 0.45 & 9.80 & 0.04$^{*}$ & \underline{97.30} & 16.22 & 0.41 & 9.83 & 0.05$^{*}$ \\
& VBReg~\cite{jiang2023vbreg} & 97.12 & 17.01 & 0.43 & 9.79 & 0.20$^{*}$ & 95.86 & 16.12 & 0.41 & 9.73 & 0.21$^{*}$ \\
\cmidrule(l){2-12}
& RANSAC10K~\cite{fischler1981random} & 12.97 & 64.58 & 1.86 & 30.63 & 1.92 & 65.05 & 56.14 & 1.48 & 24.57 & 1.95 \\
& RANSAC50K~\cite{fischler1981random} & 26.49 & 60.36 & 1.72 & 24.28 & 9.37 & 76.58 & 47.07 & 1.26 & 20.90 & 9.48 \\
& TEASER++~\cite{yang2021teaser} & 90.63 & 37.01 & 1.11 & 18.21 & \textbf{0.07} & 94.77 & 29.49 & 0.86 & 14.90 & \textbf{0.08} \\
& SC2-PCR~\cite{chen2022sc2} & 97.48 & 16.75 & 0.43 & 9.57 & 0.90 & 96.94 & 15.99 & 0.41 & 9.66 & 0.90 \\
& FastMAC~\cite{zhang2024fastmac} & 93.69 & 29.08 & 0.82 & 13.20 & 0.63 & 71.89 & 34.36 & 0.92 & 14.22 & 0.67 \\
& MAC~\cite{zhang20233d} & 97.48 & \underline{15.57} & \underline{0.41} & \underline{8.62} & 3.91 & 97.12 & \underline{13.61} & \underline{0.36} & 7.99 & 0.99 \\
& TCF~\cite{shi2024ransac} & 97.84 & 21.13 & 0.53 & 11.85 & \textbf{0.07} & 96.22 & 23.17 & 0.61 & 13.01 & 0.39 \\
& MAC++~\cite{zhang2025mac} & \underline{98.02} & 25.25 & 0.77 & 15.11 & 9.49
 & 95.86 & 16.85 & 0.45 & 9.97 & 2.86\\
& TurboReg~\cite{yan2025turboreg} & 97.48 & 16.30 & 0.45 & 8.67 & 0.03$^*$ & \textbf{97.84} & 13.95 & 0.38 & \underline{7.80} & 0.03$^*$\\
& Ours & \textbf{98.20} & \textbf{12.26} & \textbf{0.28} & \textbf{7.59} & \underline{0.12} & 96.58 & \textbf{10.78} & \textbf{0.27} & \textbf{6.65} & \underline{0.36} \\
\bottomrule
\end{tabular}
\vspace{0.2cm}

\begin{minipage}{\textwidth}
\scriptsize
\hspace{0.5cm}\footnotesize{
Note: an asterisk ($^{*}$) indicates the execution time from the GPU.
}
\end{minipage}
\end{table*}

\mypara{Comparisons with State-of-the-Art Methods}
As shown in Table~\ref{tab:compare_all}, our method achieves state-of-the-art registration accuracy (RMSE, RE and TE) across all datasets. Although the registration recall of our method is slightly lower than that of MAC~\cite{zhang20233d} and MAC++~\cite{zhang2025mac} on some datasets, it offers a significantly higher computational efficiency (e.g., 0.11 s for ours versus 3.29 s for MAC++). 
For computational time, our method outperforms all other CPU-based approaches except TCF~\cite{shi2024ransac}. Although our method is slightly slower than TCF in some scenarios, it notably surpasses TCF in key performance metrics, including registration recall and accuracy. Since TurboReg has not released its CPU implementation publicly, a direct runtime comparison is infeasible. However, based on their reported results on the 3DMatch dataset using FPFH feature (3.23 s for MAC vs. 0.41 s for TurboReg, a 7.9$\times$ speedup), and our experimental results (2.10 s for MAC vs. 0.14 s for our method, a 15$\times$ speedup), our approach likely offers a notable speed advantage over TurboReg.

\begin{table*}[htb]
    \caption{Numerical comparison of method variants for transformation estimation.}
	\label{tab:PCR_Performance}
	\setlength{\tabcolsep}{2.0pt}  
    \renewcommand{\arraystretch}{0.95} 
	\centering
	\begin{footnotesize}  
		\begin{tabular}{c | c | c | c c c c | c c | c c}  
			\toprule 
			\multirow{2}{*}{\textbf{Variant}} & \multirow{2}{*}{\textbf{Fast}} & \multirow{2}{*}{\textbf{Refine.}} & \multicolumn{4}{c|}{\textbf{Dual-Space Opt.}} & \multicolumn{2}{c|}{\textbf{3DLoMatch}} & \multicolumn{2}{c}{\textbf{KITTI}} \\
            \cmidrule(r){4-7} \cmidrule(r){8-9} \cmidrule(r){10-11}
            & & & \textbf{Anchor} & \textbf{Geo.} & \textbf{Patch} & \textbf{Whole} & \textbf{FPFH} & \textbf{FCGF} & \textbf{FPFH} & \textbf{FCGF} \\
			\midrule
			w/o Fast & & $\checkmark$ & $\checkmark$ & $\checkmark$ & & $\checkmark$ & 29.1 / \textbf{7.31} / 0.68 & 54.9 / \textbf{7.39} / 0.50 & 84.0 / \underline{12.81} / 0.78 & 91.4 / \textbf{10.47} / 0.66 \\
			w/o Refine. & $\checkmark$ & & $\checkmark$ & $\checkmark$ & & $\checkmark$ & 41.3 / 8.76 / \underline{0.11} & \underline{60.8} / 8.07 / \underline{0.15} & 97.7 / 13.51 / \textbf{0.12} & \textbf{96.8} / 15.36 / \textbf{0.36} \\
                w/o Dual-space Opt. & $\checkmark$ & $\checkmark$ &  &  & & & 32.1 / 12.04 / \textbf{0.07} & 49.2 / 11.16 / \textbf{0.11} & \textbf{98.2} / 28.22 / \textbf{0.12} & \underline{96.6} / 27.52 / 0.41 \\
            \midrule 
            w/o Anchor & $\checkmark$ & $\checkmark$ & & $\checkmark$ & & $\checkmark$ & 41.6 / 7.99 / 0.13 & 60.3 / \underline{7.74} / 0.17 & \underline{98.0} / 19.16 / \underline{0.13} & 94.8 / 17.78 / \underline{0.38} \\
            w/o Geo. & $\checkmark$ & $\checkmark$ & $\checkmark$ & & & $\checkmark$ & 37.0 / 9.51 / \underline{0.11} & 55.4 / 9.22 / \underline{0.15} & \textbf{98.2} / 16.36 / \underline{0.13} & \underline{96.6} / 13.25 / \underline{0.38} \\
            \midrule
			Patch & $\checkmark$ & $\checkmark$ & $\checkmark$ & $\checkmark$ & $\checkmark$ & & \textbf{41.9} / 8.40 / 0.17 & 59.7 / 7.90 / 0.28 & \textbf{98.2} / 14.42 / 0.15 & \underline{96.6} / 12.65 / 0.54 \\
			Whole (Ours) & $\checkmark$ & $\checkmark$ & $\checkmark$ & $\checkmark$ & & $\checkmark$ & \underline{41.7} / \underline{7.98} / \underline{0.11} & \textbf{60.9} / 7.81 / 0.16 & \textbf{98.2} / \textbf{12.26} / \underline{0.12} & \underline{96.6} / \underline{10.78} / \textbf{0.36} \\
			\bottomrule
		\end{tabular}
	\end{footnotesize}
\vspace{0.15cm}

\begin{minipage}{\textwidth}
\scriptsize
\hspace{0.5cm}\footnotesize{Note: the values in the table represent: RR(\%) $^{\uparrow}$ / RMSE(cm) $^{\downarrow}$ / Time(s) $^{\downarrow}$ by different variants.}
\end{minipage}
\end{table*}

\begin{table}[t]
\centering 
\caption{Numerical comparison of method variants for outlier removal.}
\label{tab:outlier_removal}
\setlength{\tabcolsep}{1.5pt}
\scriptsize
\begin{tabular}{@{}c|ccc|cc|cc|cc@{}}
\toprule
\multirow{2}{*}{\textbf{Variant}} & \multicolumn{3}{c|}{\textbf{Fast Filtering}} & \multicolumn{2}{c|}{\textbf{Refinement}} & \multicolumn{2}{c|}{\textbf{3DLoMatch}} & \multicolumn{2}{c}{\textbf{KITTI}} \\
\cmidrule(lr){2-4} \cmidrule(lr){5-6} \cmidrule(lr){7-8} \cmidrule(lr){9-10}
 & $D_L$ & $D_N$ & \textbf{Symm.} & \textbf{Random} & \textbf{Weighted} & \textbf{FPFH} & \textbf{FCGF} & \textbf{FPFH} & \textbf{FCGF} \\
\midrule
1 & $\checkmark$ &  &  &  &  & 33.00 & 48.37 & 45.65 & 44.10\\
2 & $\checkmark$ & $\checkmark$ &  &  &  & 33.55 & 48.64 & 46.17 & 44.75\\
3 & $\checkmark$ & $\checkmark$ & $\checkmark$ &  &  & 36.00 & 52.97 & 46.13 & 45.04\\
4 & $\checkmark$ & $\checkmark$ & $\checkmark$ & $\checkmark$ &  & \underline{37.10} & \underline{53.85} & \underline{84.05} & \underline{68.07}\\
Ours & $\checkmark$ & $\checkmark$ & $\checkmark$ &  & $\checkmark$ & \textbf{37.22} & \textbf{53.87} & \textbf{85.52} & \textbf{68.10}\\
\bottomrule
\end{tabular}

\begin{minipage}{\columnwidth}
\scriptsize
\footnotesize{Note: the values in the table represent the ground truth inlier ratio (\%) $^{\uparrow}$ in the filtered correspondence set by different variants.}
\end{minipage}
\end{table}

\mypara{Effectiveness of Outlier Removal}
To verify the effectiveness of each module described in Sec.~\ref{sec:filter-feature-corr} for removing outliers, we tested the following variants:
\begin{enumerate}
    \item We only use the length consistency constraint in Eq.~\eqref{eq:length-consistent} and ignore the tangential distance $D_N$ in Eq.~\eqref{eq:normal-consistent} to define the one-point-based consensus set;
    \item We use length consistency and tangential distance to define the consistency set according to Eqs.~\eqref{eq:init-one-point-consensus} and \eqref{eq:one-point-consensus-set}, without applying the symmetry check;
    \item We use the complete module of the 1-point RANSAC-based approach without refinement;
    \item We use the complete module of the 1-point RANSAC-based approach. During the refinement process, the classic RANSAC algorithm is used. That is, there is no probability-based weighted sampling strategy.
\end{enumerate}  
As shown in Table~\ref{tab:outlier_removal}, the progressive introduction of length consistency, tangential distance consistency, and symmetry check leads to a steady improvement in the inlier ratio. 
Notably, the refinement module (in \textit{Variant 4} and \textit{Ours}) substantially enhances performance in outdoor scenarios. Ultimately, by incorporating the probability-based weighted sampling strategy inspired by~\cite{piedade2023bansac}, our full method achieves the best overall performance among all variants.

\mypara{Effectiveness of Transformation Estimation}
We further tested the influence of each module in our method on the registration accuracy. We tested the
following variants: 
\begin{itemize}
    \item \textit{w/o Fast \& w/o Refine. \& w/o Dual-space Opt.}:  Our method ignores the 1-point RANSAC-based fast filtering, the 3-point RANSAC-based refinement, or the dual-space optimization, respectively.
    \item \textit{w/o Anchor \& w/o Geo.: } Our robust estimation algorithm ignores the filtered feature-based correspondences (anchor points) or correspondences enhanced by spatial geometric information. 
    \item \textit{Patch \& Whole (Ours): } For the geometry-related correspondences, we can also establish them patch by patch by finding the closest point of $\mathcal{P}_{\mathbf{c}_j}^t$ for each point in $\mathcal{P}_{\mathbf{c}_j}^s$ rather than aggregating these neighborhoods as a whole. 
\end{itemize}
As shown in Table~\ref{tab:PCR_Performance}, 
the introduction of fast filtering brings significant accuracy improvements across all datasets (comparing rows 1 and 7), while the refinement algorithm further enhances the registration results (comparing rows 2 and 7). Our method outperforms the variant without dual-space optimization (row 3), highlighting the critical importance of this strategy. The synergy between feature-based anchor points and geometry-based correspondences is essential in dual-space optimization, as removing either component (rows 4 and 5) leads to reduced accuracy. Notably, on the 3DLoMatch dataset with FCGF features, the complete method achieves 60.9\% accuracy, while removing geometry-based correspondences and anchor points reduces accuracy to 55.4\% and 60.3\%.
Furthermore, the whole geometry-based correspondence establishment outperforms patch-wise processing (comparing rows 7 and 6), demonstrating lower RMSE and better overall performance.

\section{Conclusion}
In this paper, we presented an efficient and robust rigid registration framework integrating novel correspondence filtering and dual-space optimization. Correspondences are initially filtered using a 1-point RANSAC and a probability-weighted 3-point RANSAC refinement. These points then serve as anchors to construct geometric proxies, yielding consistent candidates. Finally, a dual-space optimization algorithm with adaptive weights estimates the optimal transformation. Extensive experiments have demonstrated our method's superior accuracy and efficiency.
Nevertheless, it struggles with extremely low inlier ratios, as RANSAC-based methods inherently assume the inlier set is larger than any outlier cluster. Incorporating better initial correspondences could further improve registration accuracy.

\mypara{Acknowledgments}
This research was supported by the National Natural Science Foundation of China (No.62272433), Anhui Provincial Natural Science Foundation (No.2508085ZD011) and the Fundamental Research Funds for the Central Universities.
{
    \small
    \bibliographystyle{ieeenat_fullname}
    \bibliography{main}
}

\clearpage
\setcounter{page}{1}
\maketitlesupplementary
\appendix

\section{Technical Details}
\label{sec: Technical Details}
\mypara{Disambiguating symmetry}
In Sec.~\ref{sec:consensus-defination}, we check whether the established set of corresponding points exhibits symmetry; if it does, it will not be a reliable set of correspondences.  Specifically, we compute the orthogonal matrix 
\begin{equation}
\label{eq:solve-M}
\begin{aligned}
\mathbf{M}^* =& \argmin_{\mathbf{M}\in\mathbb{R}^{3\times 3}} \sum_{\mathbf{c}_i\in\mathcal{I}(\mathbf{c}_j)} \|\mathbf{M}\mathbf{v}_i + \mathbf{t}_M - \mathbf{u}_i\|^2,\\
& \text{s.t.~~} \mathbf{M}^T\mathbf{M} = \mathbf{I},
\end{aligned}
\end{equation}
where $\mathbf{I}\in\mathbb{R}^{3\times 3}$ is the identity matrix. $\mathbf{t}_M$ is the translation vector, which we do not need to solve here. Similar to solving the rotation matrix~\cite{RigidSVD}, we can compute Singular Value Decomposition (SVD) 
\[
 \mathbf{U_s}\Sigma \mathbf{V_s}^T = \sum_{\mathbf{c}_i\in\mathcal{I}(\mathbf{c}_j)} (\mathbf{v}_i-\overline{\mathbf{v}})(\mathbf{u}_i-\overline{\mathbf{u}})^T, 
\]
where 
\[
\overline{\mathbf{v}}=\sum_{\mathbf{c}_i\in\mathcal{I}(\mathbf{c}_j)}\mathbf{v}_i/|\mathcal{I}(\mathbf{c}_j)|\text{~and~}\overline{\mathbf{u}}=\sum_{\mathbf{c}_i\in\mathcal{I}(\mathbf{c}_j)}\mathbf{u}_i/|\mathcal{I}(\mathbf{c}_j)|.
\]
Then we can compute $\mathbf{M}^*=\mathbf{V_s}\mathbf{U}_s^T$.  When $\det(\mathbf{M}^*)=1$, $\mathbf{M}^*$ is a rotation matrix, but when $\det(\mathbf{M}^*)=-1$, $\{\mathbf{v}_i\}_{\mathbf{c}_i\in\mathcal{I}(\mathbf{c}_j)}$ and $\{\mathbf{u}_i\}_{\mathbf{c}_i\in\mathcal{I}(\mathbf{c}_j)}$ are symmetric about a plane in 3D space. Therefore, we set $\mathcal{I}(\mathbf{c}_j)=\emptyset$ if $\det(\mathbf{M}^*)=-1$ to eliminate the effect of symmetry.

\mypara{Algorithmic pipeline of 1-point RANSAC}
We also present the algorithmic pipeline of our proposed 1-point RANSAC for fast filtering of feature correspondences, as detailed in Alg.~\ref{alg:one-point-ransac}.

\begin{algorithm}[t]
\caption{One-Point RANSAC Algorithm}
\label{alg:one-point-ransac}
\KwIn{$\mathcal{V},\mathcal{N}_s$: the source points and normals\;
~~~~$\mathcal{U},\mathcal{N}_t$: the target points and normals\;
~~~~$\mathcal{C}_0$: initial correspondence set\; 
~~~~$\{s_{\mathbf{c}_j}\}$: confidence scores of correspondences $\{\mathbf{c}_j\}$\;
~~~~$N_{\text{I}}$: the number of estimated inliers in $\mathcal{C}_0$\; 
} 
\KwOut{Filtered correspondence set $\mathcal{C}_{\text{I}}$;}
Set $s_{\mathbf{c}_j}^{(0)}=0$ for $\forall \mathbf{c}_j\in\mathcal{C}_0$\; 
$K_{\text{I}}^{(0)}=\infty$;~~$N_{\text{I}}^{(0)}=1$; ~~$k=1$\;
\While{$k<K_{\text{I}}^{(k-1)}$}{
Randomly sample a correspondence $\mathbf{c}_j^{(k)}\in\mathcal{C}_0$\;
Compute the consensus set $\mathcal{I}(\mathbf{c}_j^{(k)})$ with Eq.~\eqref{eq:one-point-consensus-set}\;
Solve orthogonal matrix $\mathbf{M}^*$ by~\eqref{eq:solve-M}\;
\If{$\det(\mathbf{M}^*)=-1$}{$\mathcal{I}(\mathbf{c}_j^{(k)})=\emptyset$;}
\If{$|\mathcal{I}(\mathbf{c}_j^{(k)})|>\alpha\cdot N_{\text{I}}^{(k-1)}$}{
\For{$\forall \mathbf{c}_i\in\mathcal{I}(\mathbf{c}_j^{(k)})$}{
$s_{\mathbf{c}_i}^{(k+1)}=s_{\mathbf{c}_i}^{(k-1)}+1$\;
}
}
$N_{\text{I}}^{(k)}=\max(N_{\text{I}}^{(k-1)},~ |\mathcal{I}(\mathbf{c}_j^{(k)})|)$\;
Compute $K_{\text{I}}^{(k)}$ with Eq.~\eqref{eq:one-point-ransac-termination}\;
$k = k+1$\;
}
Compute $\mathcal{C}_{\text{I}}=\mathcal{I}(\mathbf{c}_*)$ by Eq.~\eqref{eq:best-C-coarse}\;
\end{algorithm}

\mypara{Details of 3-point RANSAC-based refinement}
\label{sec:supp-refinement}
We denote the initial inlier probability of each correspondence $\mathbf{c}_j\in\mathcal{C}_{\text{I}}$ as $p_{\mathbf{c}_j}$, best transformation as $(\mathbf{R}_{\text{II}}^*,\mathbf{t}_{\text{II}}^*)$, the estimated inlier set as $\mathcal{C}_{\text{II}}$. Initially, we set $\mathcal{C}_{\text{II}}=\emptyset$, $p_{\mathbf{c}_j}=0.5$ for any $\mathbf{c}_j\in\mathcal{C}_{\text{I}}$, and perform the algorithm at the $k$-th iteration as follows: 
\begin{itemize}
    \item Randomly take a sample $\mathbf{S}^{(k)}=\{\mathbf{c}_i, \mathbf{c}_j, \mathbf{c}_l\in\mathcal{C}_{\text{I}}\}$ according to the probabilities.
    \item Compute the rigid transformation $(\mathbf{R}_{\mathbf{S}^{(k)}},\mathbf{t}_{\mathbf{S}^{(k)}})$ so that 
    \begin{equation}
    \label{eq:solve-RT-BANSAC}
    \begin{aligned}
    (\mathbf{R}_{\mathbf{S}^{(k)}},\mathbf{t}_{\mathbf{S}^{(k)}}) &= \argmin_{\mathbf{R},\mathbf{t}} \sum_{\mathbf{c}\in\{\mathbf{c}_i, \mathbf{c}_j, \mathbf{c}_l\}} \|\mathbf{R}\mathbf{v} + \mathbf{t} - \mathbf{u}\|^2,\\
    & \text{s.t.~~} \mathbf{R}^T\mathbf{R} = \mathbf{I} \text{~and~} \det{(\mathbf{R})}=1.
    \end{aligned}
    \end{equation}
    It can be solved in closed form via SVD~\cite{RigidSVD}. 
    \item Evaluate $(\mathbf{R}_{\mathbf{S}^{(k)}},\mathbf{t}_{\mathbf{S}^{(k)}})$. We regard correspondences that meet $\|\mathbf{R}_{\mathbf{S}^{(k)}}\mathbf{v}+\mathbf{t}_{\mathbf{S}^{(k)}}-\mathbf{u}\|<\gamma$ as inliers $\mathcal{I}_{\text{II}}^{(k)}$ and others as outliers. 
    If $|\mathcal{I}_{\text{II}}^{(k)}|>|\mathcal{C}_{\text{II}}|$, we set $\mathcal{C}_{\text{II}} = \mathcal{I}_{\text{II}}^{(k)}$ and $(\mathbf{R}_{\text{II}}^*,\mathbf{t}_{\text{II}}^*)=(\mathbf{R}_{\mathbf{S}^{(k)}},\mathbf{t}_{\mathbf{S}^{(k)}})$. 
    \item Update probabilities according to current and previous inlier/outlier classifications. 
\end{itemize}
For more details on probability-based sampling and probability updating, please refer to~\cite{piedade2023bansac}. We adopt a termination criterion similar to that of coarse filtering, i.e., dynamically updating the maximum number of iterations 
\begin{equation}
\label{eq:bansac-termination}
K_{\text{II}}^{(k)} = \left\lceil\frac{\log (1-\lambda)}{\log(1-(|\mathcal{I}_{\text{II}}^{(k)}|/|\mathcal{C}|)^3)}\right\rceil. 
\end{equation}
The best transformation $(\mathbf{R}_{\text{II}}^*,\mathbf{t}_{\text{II}}^*)$ and the inlier set $\mathcal{C}_{\text{II}}$ will be obtained to perform robust transformation estimation. 

\mypara{Implementation details}
For the 3DMatch, 3DLoMatch, and KITTI datasets, we generate initial correspondences using Fast Point Feature Histogram (FPFH) and Fully Convolutional Geometric Features (FCGF) as in MAC~\cite{zhang20233d}. 
Normals of point clouds are computed using PyMeshLab's point cloud normal estimation algorithm. 
All point clouds are downsampled by PCL's VoxelGrid method. 
The length consensus parameter $\tau$ in Sec.~\ref{sec:consensus-defination}, downsampling parameter, and neighborhood radius $\beta$ in Sec.~\ref{sec:Geometric-Proxies} are all set to multiples of the point cloud resolution (3$\times$, 5$\times$, 50$\times$). The inlier threshold $\gamma$ in Sec.~\ref{sec: Technical Details} follows the configuration in~\cite{zhang20233d}, while the normal consensus parameter $\nu=2\gamma$ in Sec.~\ref{sec:consensus-defination}. The parameter $\alpha$ in Sec.~\ref{sec:Searching-Optimal-Consensus-Set} is configured as 0.2 for 3DMatch, 0.95 for 3DLoMatch and KITTI with FCGF feature, and 0.9 for KITTI with FPFH feature. The parameter $\lambda$ in Sec.~\ref{sec:Optimization-Formulation} is set to 0.05 for 3DMatch/3DLoMatch and 1.0 for KITTI.
To evaluate the robustness of our parameters, we conducted a sensitivity analysis on the parameters used in our method in Sec.~\ref{sec:Parameter_Sensitivity_Analysis}.

\begin{figure*}[!htb]
  \centering
  \hspace*{-1.8em}
  \includegraphics[width=1.05\textwidth]{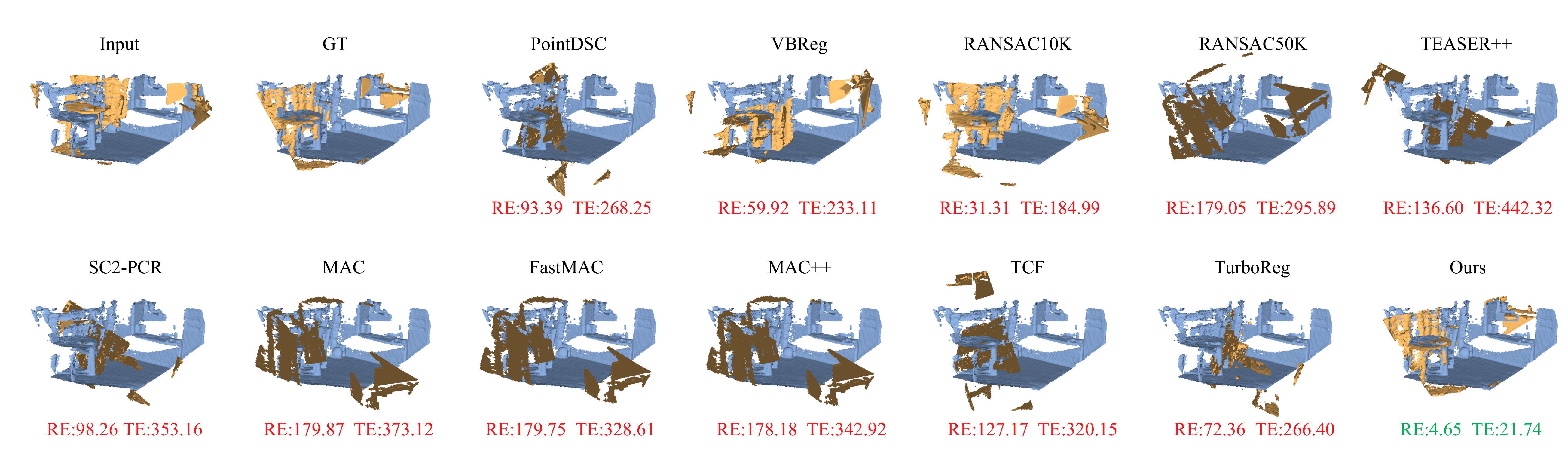}
  \hspace*{-1em}
  \includegraphics[width=1.05\textwidth]{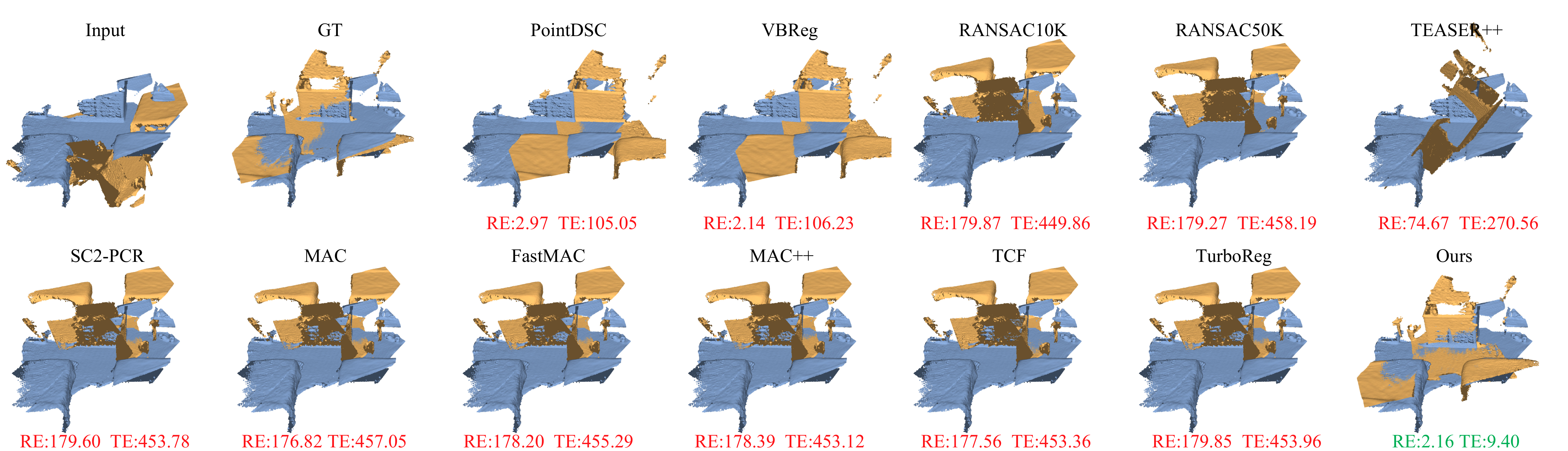}

  \caption{Qualitative comparison on 3DMatch. The yellow and blue point clouds represent the source and target point clouds, respectively. The first two rows correspond to FPFH-based examples with a ground truth inlier ratio of 1.55\%, while the latter two rows demonstrate FCGF-based cases exhibiting a ground truth inlier ratio of 6.39\%. Rotation error (RE) and translation error (TE) are marked, with red and green representing failed and successful registration, respectively.} 
  \label{fig:3dMatch} 
\end{figure*}

\mypara{Evaluation Criteria}
We evaluate the performance on all datasets using registration recall (RR), root
mean square error (RMSE), rotation error (RE), and translation error (TE), refer to~\cite{choy2020deep,zhang20233d}.
The registration success rate is the ratio of successful registration cases to all cases.
For 3DMatch, 3DLoMatch datasets, a registration is considered successful when $\mathrm{RE} < 15^{\circ}$ and $\mathrm{TE} < 30\,\mathrm{cm}$. For the KITTI dataset, the success criteria require $\mathrm{RE} < 5^{\circ}$ and $\mathrm{TE} < 60\,\mathrm{cm}$. 
We additionally record the computational time of all methods to demonstrate our approach's efficiency.

\section{More Analysis and Visualization}

\subsection{Analysis and Visualization of Comparisons}
\begin{figure*}[!htb]
  \centering
  \hspace*{-1.8em}
  \includegraphics[width=1.05\textwidth]{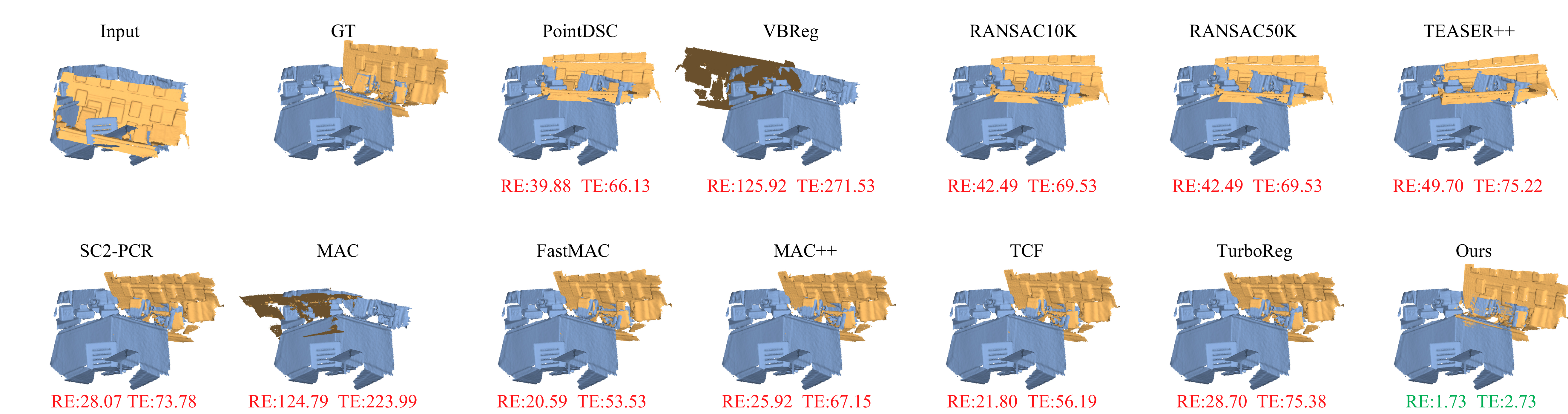}
  
  \vspace{0.2cm}
  
  \hspace*{-2.2em}
  \includegraphics[width=1.05\textwidth]{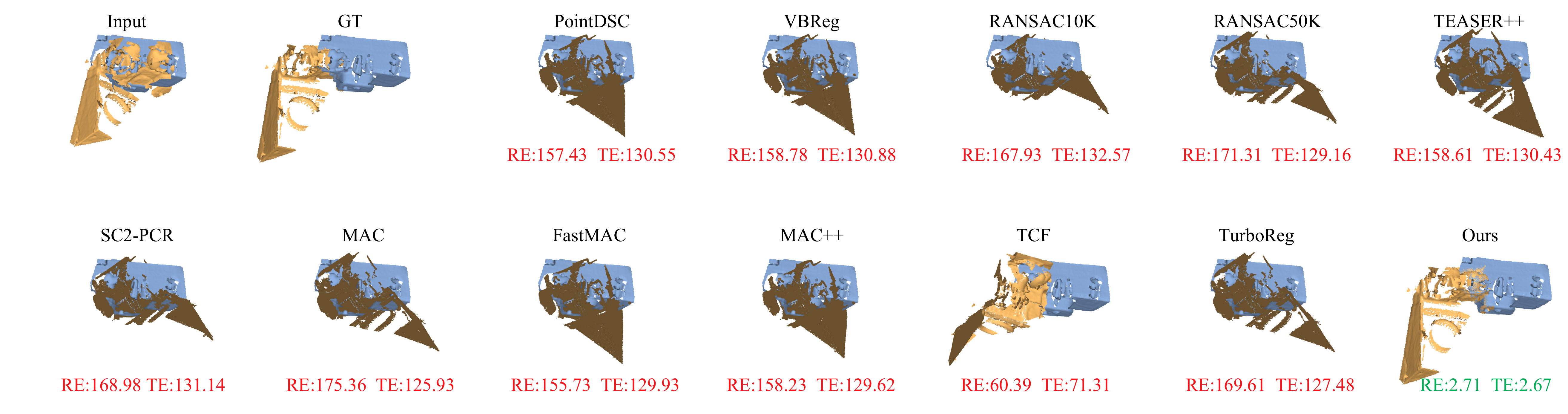}

  \caption{Qualitative comparison on 3DLoMatch. The yellow and blue point clouds represent the source and target point clouds, respectively. The first two rows correspond to FPFH-based examples with a ground truth inlier ratio of 1.90\%, while the latter two rows demonstrate FCGF-based cases exhibiting a ground truth inlier ratio of 7.32\%. Rotation error (RE) and translation error (TE) are marked, with red and green representing failed and successful registration, respectively.} 
  \label{fig:3dLoMatch} 
\end{figure*}

\begin{figure*}[!htb]
  \centering
  \includegraphics[width=1\textwidth]{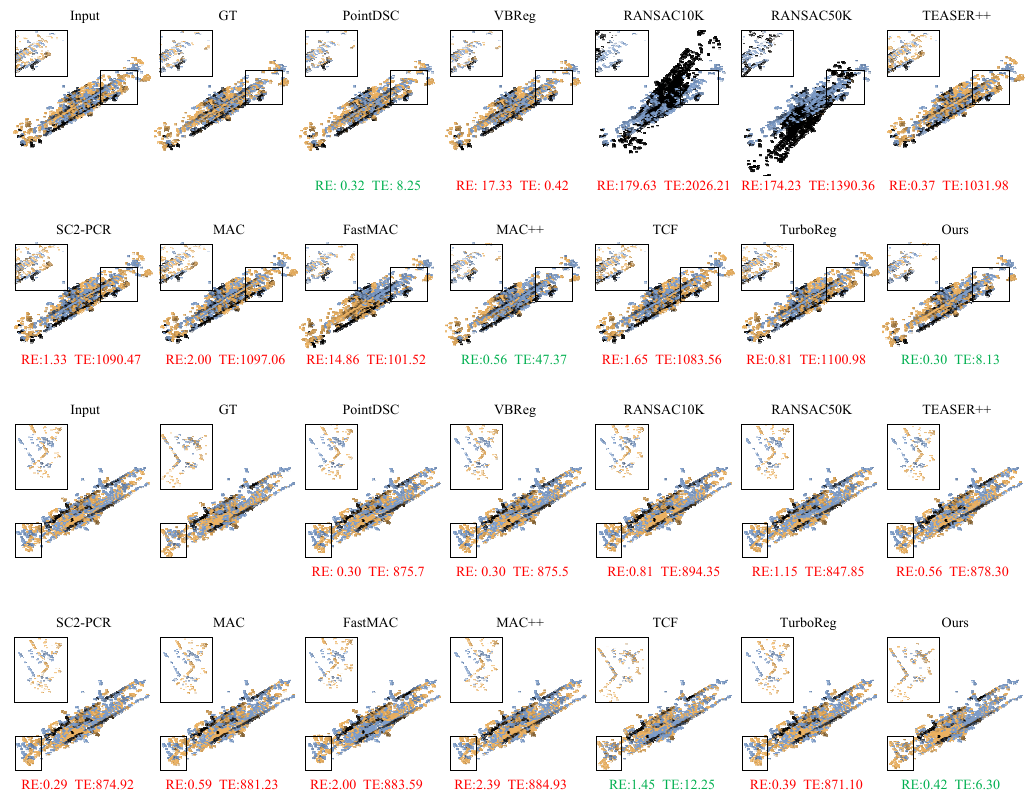}

  \caption{Qualitative comparison on KITTI. The yellow and blue point clouds represent the source and target point clouds, respectively. The first two rows correspond to FPFH-based examples with a ground truth inlier ratio of 1.32\%, while the latter two rows demonstrate FCGF-based cases exhibiting a ground truth inlier ratio of 4.12\%. Rotation error (RE) and translation error (TE) are marked, with red and green representing failed and successful registration, respectively.} 
  \label{fig:kitti} 
\end{figure*}

In the accompanying \textbf{video demo}, we visualize the entire process of the proposed method and some comparison results.
In Figures~\ref{fig:3dMatch}, \ref{fig:3dLoMatch}, and \ref{fig:kitti}, we also visualize several cases of the proposed method with the state-of-the-art methods from the 3DMatch dataset, 3DLoMatch dataset and KITTI dataset. 
In the following, we further analyze the comparison results in more detail. 

\mypara{3DMatch Dataset}
On the 3DMatch dataset, our method demonstrates comprehensive performance advantages, achieving significant breakthroughs in precision, registration rate, and speed. As shown in Table~\ref{tab:compare_all} and Fig.~\ref{fig:3dMatch}, when using the FPFH feature, we not only realized the lowest rotation error (RE = 1.75°) and translation error (TE = 6.19 cm) to date, significantly surpassing the precision levels of the previously best methods VBReg and MAC++, but also increased the registration recall (RR) to 84.41\%, setting a new high for this feature. More impressively, under the FCGF feature: our method similarly achieved exceptional precision metrics, achieving the lowest RE (1.65°) and TE (6.13 cm), far ahead of the previously best method FastMAC; in terms of registration recall, we reached an excellent level of 93.65\%, successfully matching the top record held by the benchmark method MAC in this field. 
In terms of runtime efficiency, the method also demonstrates commendable performance: when using the FPFH feature, it ranks second among all CPU-based methods with a processing time of 0.14 seconds, only 0.06 seconds behind the fastest method, TCF, while achieving significant improvements in registration success rate and accuracy. Under the FCGF feature, the method's speed even rises to first place (0.45 seconds), outperforming the previously fastest TCF by 0.35 seconds.
These results fully demonstrate that our method has achieved a perfect synergy of precision, success rate, and speed on 3DMatch, setting a new benchmark for overall performance. 

\mypara{3DLoMatch Dataset}
Facing the highly challenging 3DLoMatch dataset, our method achieves outstanding precision with several key metrics setting new state-of-the-art results, while maintaining top-tier runtime efficiency. As shown in Table~\ref{tab:compare_all} and Fig.~\ref{fig:3dLoMatch}, under the FPFH feature, we attained the lowest RMSE (7.98 cm), rotation error (RE = 3.08°), and translation error (TE = 9.36 cm), demonstrating significant precision advantages. Our registration recall (RR = 41.66\%) is highly competitive, only surpassed by MAC++ (RR = 44.58\%). Under the FCGF feature, our method again secured the lowest RMSE (7.81 cm), RE (3.06°), and TE (9.61 cm), significantly outperforming other methods (e.g., SC2-PCR: RE = 3.75°, TE = 10.44 cm; TurboReg: RE = 3.76°, TE = 10.35 cm). Our registration recall (RR = 60.86\%) is also excellent, very close to the top-performing MAC++ (RR = 61.09\%), with a minimal gap of less than 0.23\%. Crucially, this superior precision is achieved with exceptional efficiency
(among CPU-based methods): under FPFH, our time (0.11 s) is second only to the fastest TCF (0.06 s), while significantly outperforming TCF in precision (e.g., the registration recall leads by approximately 9.09\%). Under FCGF, our time consumption (0.16 s) remains second fastest, only slightly slower than TCF (0.12 s), but significantly faster than other high-precision methods (e.g., 12.3$\times$ faster than MAC's 1.97 s and 23.25$\times$ faster than MAC++'s 3.72 s). This robustly confirms that even on the demanding 3DLoMatch benchmark, our method delivers leading or near-leading precision and registration rates with top-tier efficiency, showcasing exceptional robustness and practicality. 

\mypara{KITTI Dataset}
On the KITTI dataset, our method has established a clear leading advantage in multiple core indicators, demonstrating strong comprehensive performance. The numerical results and visualization results are shown in Table~\ref{tab:compare_all} and Fig.~\ref{fig:kitti}. Under the FPFH feature, we achieved the highest registration recall (RR = 98.20\%) and simultaneously obtained the lowest RMSE (12.26 cm), rotation error (RE = 0.28°), and translation error (TE = 7.59 cm), with a precision (RE, TE) that achieved a considerable improvement compared to the previously best-performing method MAC (RMSE = 15.57 cm, RE = 0.41°, TE = 8.62 cm). Under the FCGF feature representation, the precision superiority of our method remains equally pronounced; we obtained the lowest RMSE (10.78 cm), RE (0.27°), and TE (6.65 cm), comprehensively and significantly leading other competitors (e.g., MAC: RMSE = 13.61 cm, RE = 0.36°, TE = 7.99 cm); the registration recall also reached an excellent level of 96.58\%, almost on par with the best method TurboReg (97.84\%). In the crucial aspect of runtime speed, the method's performance is equally noteworthy (among CPU-based methods): under the FPFH feature, the time consumption is only 0.12 s, although 0.05 s slower than the fastest TCF (0.07 s) and TEASER++ (0.07 s), we have achieved significant advantages in registration rate and precision; under the FCGF feature, the speed of 0.36 s is not only highly competitive but also more than twice as fast as the MAC (0.99 s) with similar precision performance, significantly enhancing practicality. Overall, our method has excellently balanced high precision, high registration rate, and high efficiency on KITTI, with outstanding comprehensive performance.

\subsection{Runtime Breakdown}
Additionally, we provide a runtime breakdown of our method across all datasets, as shown in Table~\ref{tab:time_analysis}.

\begin{table}[htbp]
\centering
\caption{Average computation times (s).}
\label{tab:time_analysis}
\setlength{\tabcolsep}{3.2pt}  
\small
\begin{tabular}{l|cccc}
\toprule
Dataset & Fast & Refine. & Dual-space Opt. & All \\
\midrule
3DMatch+FPFH & 0.06 & 4.09E-05 & 0.07 & 0.14 \\
3DMatch+FCGF & 0.36 & 5.12E-05 & 0.09 & 0.45 \\
3DLoMatch+FPFH & 0.06 & 4.37E-05 & 0.06 & 0.11 \\
3DLoMatch+FCGF & 0.09 & 4.21E-05 & 0.07 & 0.16 \\
KITTI+FPFH & 0.10 & 4.09E-04 & 1.89E-02 & 0.12 \\
KITTI+FCGF & 0.33 & 1.33E-03 & 2.48E-02 & 0.36 \\
\bottomrule
\end{tabular}
\end{table}

\subsection{Robustness to Varying Inlier Ratios}
\begin{figure}[!htb] 
    \centering
\includegraphics[width=1.0\columnwidth]{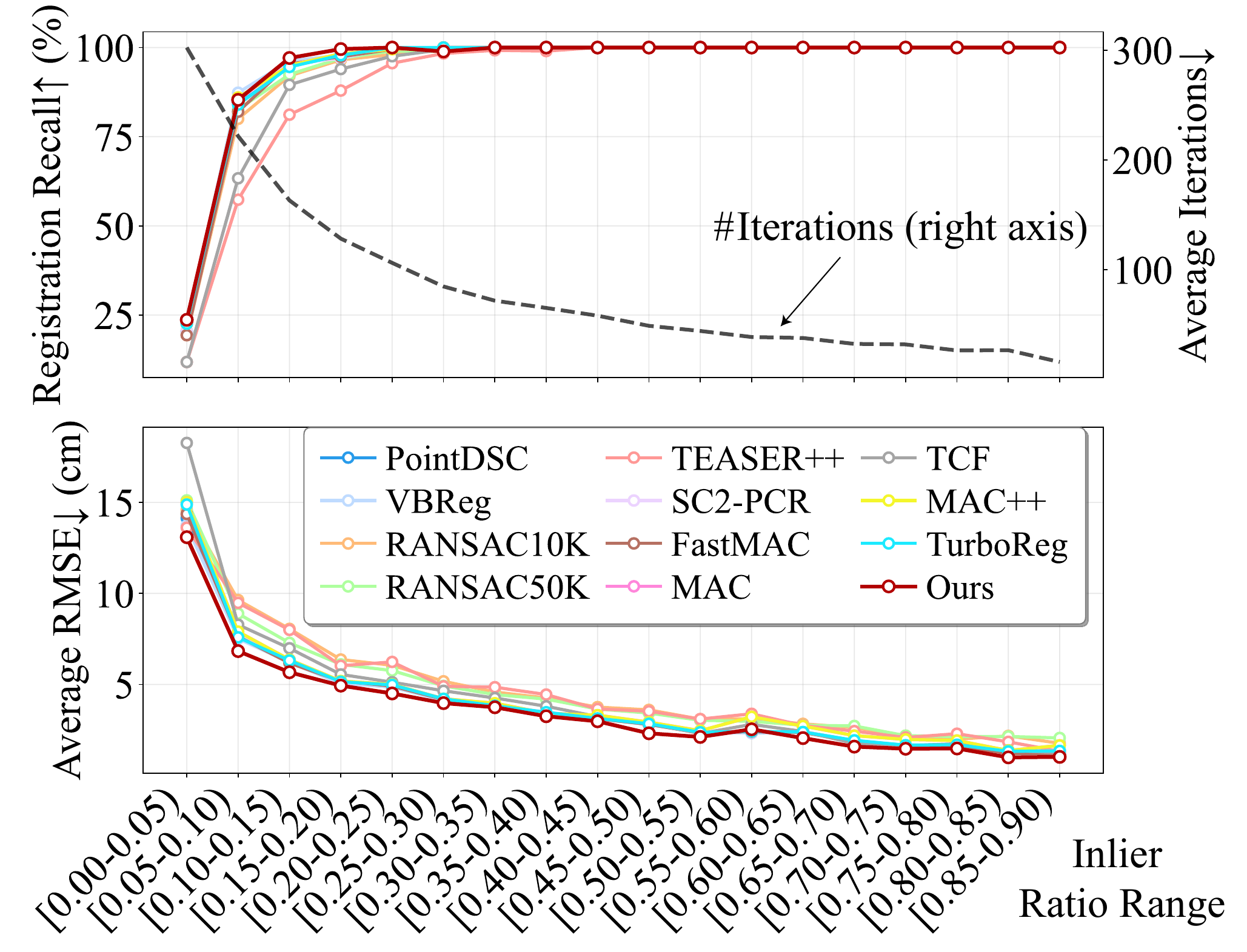}
    \caption{\footnotesize{Statistics across different inlier ratio ranges.}}
    \label{fig:rebuttal_curve_figure} 
\end{figure}

To more intuitively demonstrate our method's robustness, we plot curves of registration recall and RMSE under varying inlier ratios on the 3DMatch dataset using FCGF features, and compare them with other methods, as shown in Figure~\ref{fig:rebuttal_curve_figure}. By comparing these trends with existing methods, we further validate the robustness and effectiveness of our approach. Furthermore, we also plot the curve of the iteration count (i.e., the number of iterations in the fast filtering process) to illustrate the efficacy of our method. 

\subsection{Robustness of Normal-based Filtering}
To address the inherent sign ambiguity in normal estimation, we propose using the bijective point-to-plane distance as our normal-based filtering strategy. In contrast, a commonly used baseline is the standard angular difference. Given two correspondences $\mathbf{c}_i=(\mathbf{v}_i,\mathbf{u}_i)$ and $\mathbf{c}_j=(\mathbf{v}_j,\mathbf{u}_j)$ alongside their normals $\mathbf{n}_i^s, \mathbf{n}_i^t, \mathbf{n}_j^s, \mathbf{n}_j^t$, the baseline evaluates the difference between $\angle(\mathbf{n}_i^s,\mathbf{n}_j^s)$ and $\angle(\mathbf{n}_i^t,\mathbf{n}_j^t)$, rejecting the pair if the deviation exceeds a predefined threshold. Evidently, our strategy is inherently more robust to normal inversions. The numerical comparisons in Table~\ref{tab:supp_normal_compare} further validate the superior effectiveness of our approach.

\begin{table}[htbp]
\centering
\caption{Comparison of filtering effects in the fast filtering stage.}
\small
\label{tab:supp_normal_compare}
\begin{tabular}{l|cc}
\toprule
Dataset & \(D_N\) (Ours) & Angular Diff. \\
\midrule
3DLoMatch+FPFH & \textbf{36.00} & 35.10 \\
3DLoMatch+FCGF & \textbf{52.97} & 52.52 \\
KITTI+FPFH & \textbf{46.13} & 45.55 \\
KITTI+FCGF & \textbf{45.04} & 44.34 \\
\bottomrule
\end{tabular}
\begin{minipage}{\columnwidth}
\scriptsize
\footnotesize{Note: the values in the table represent the ground truth inlier ratio (\%) $^{\uparrow}$ in the filtered correspondence set by different normal filtering strategies.}
\end{minipage}
\end{table}

To further validate our method's robustness to normal estimation parameters, we vary the neighbor count ($K$) used for normal computation to 10 and 30, deviating from the default value of 20. We also substitute PyMeshLab with the Open3D library~\cite{Zhou2018open3d} to test its adaptability to different software tools. As reported in Table~\ref{Tab:supp_normal_computing}, our method exhibits only minor performance variations, confirming its strong resilience to the underlying normal estimation process.

\begin{table}[htbp]
\centering
\caption{Performance comparisons with different normal computing.}
\setlength{\tabcolsep}{5pt}  
\small
\label{Tab:supp_normal_computing}
\begin{tabular}{l|ccccc}
\toprule
Method & RR$\uparrow$ & RMSE$\downarrow$ & RE$\downarrow$ & TE$\downarrow$ & time$\downarrow$ \\
\midrule
Open3D ($K$=20) & 93.22 & \textbf{4.39} & \textbf{1.62} & \textbf{6.10} & \underline{0.45} \\
$K$=10 & \underline{93.35} & 4.43 & \underline{1.65} & 6.15 & \textbf{0.42} \\
$K$=30 & 93.16 & \underline{4.41} & \underline{1.65} & \textbf{6.10} & 0.48 \\
$K$=20 (ours) & \textbf{93.65} & 4.44 & \underline{1.65} & \underline{6.13} & \underline{0.45} \\
\bottomrule
\end{tabular}
\begin{minipage}{\columnwidth}
\scriptsize
\footnotesize{Note: RR(\%), RMSE(cm), RE($^\circ$), TE(cm), time(s).}
\end{minipage}
\end{table}

\subsection{Sensitivity Analysis of Parameters}
\label{sec:Parameter_Sensitivity_Analysis}
This section conducts a sensitivity analysis on the key parameters in our method to verify its robustness. We systematically evaluate five parameters on the 3DMatch dataset (using FCGF features): $\alpha$ in Sec.~\ref{sec:Searching-Optimal-Consensus-Set}, the downsampling parameter, $\beta$ in Sec.~\ref{sec:Geometric-Proxies}, $|\widehat{\mathcal{C}}_{\text{II}}|/|\mathcal{C}_{\text{II}}|$, and $\lambda$ in Sec.~\ref{sec:Optimization-Formulation}. Experimental results demonstrate that although we select different parameter values for various datasets (indoor/outdoor, high/low overlap) to achieve optimal performance, our method maintains stable and excellent performance across a wide range of parameter values. This demonstrates that the method is not sensitive to parameter choices and possesses good practical utility.

\begin{figure}[htbp]
    \centering
    
    \subfloat[$\alpha$ \label{fig:5a}]{
        \includegraphics[width=0.48\textwidth]{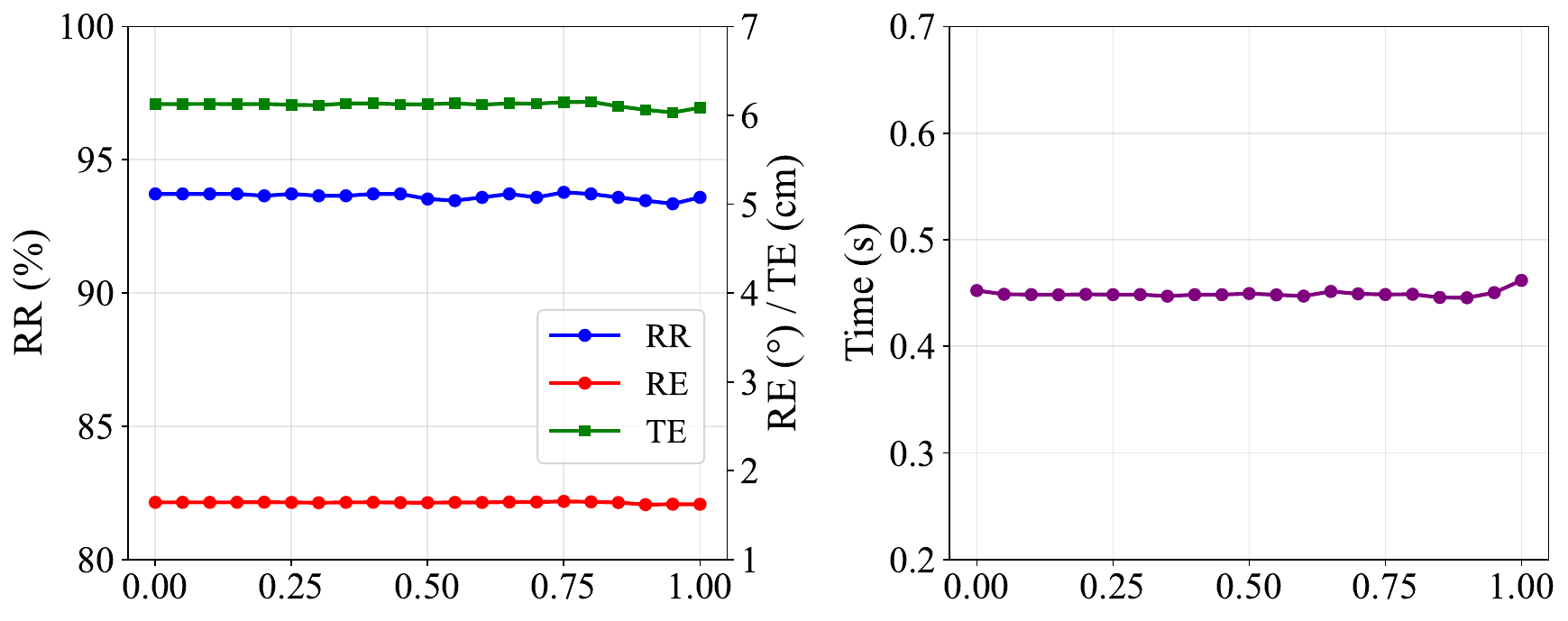}
    }
    \vspace{0.1cm}
    \subfloat[downsampling parameter \label{fig:5b}]{
        \includegraphics[width=0.48\textwidth]{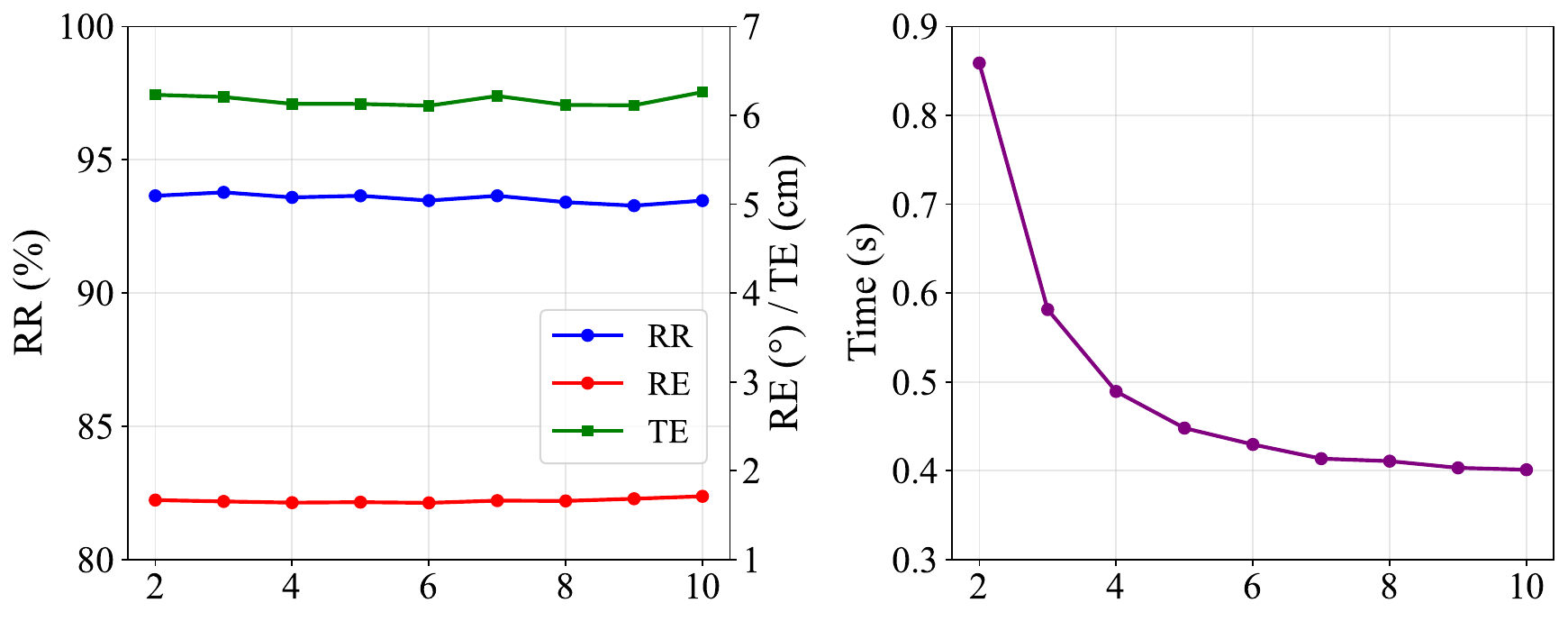}
    }
    \vspace{0.1cm}
    \subfloat[ $\beta$ \label{fig:5c}]{
        \includegraphics[width=0.48\textwidth]{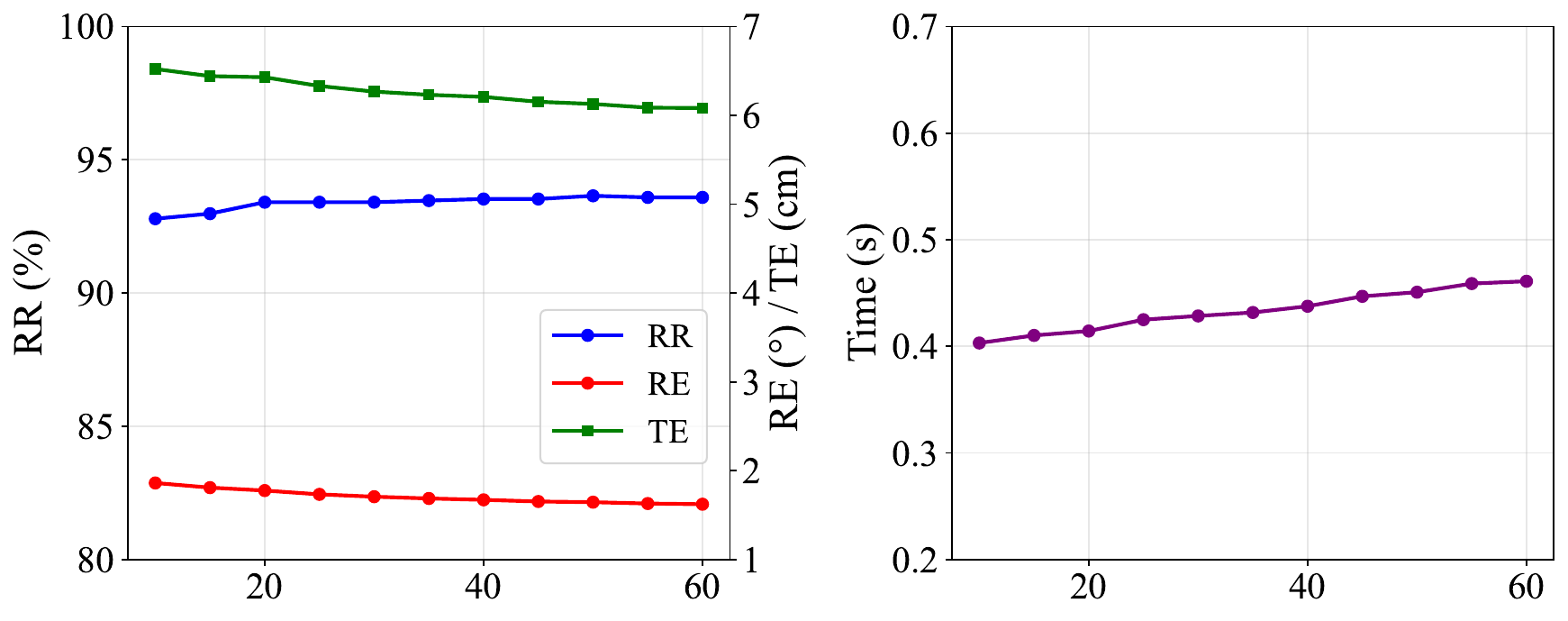}
    }
    \vspace{0.1cm}    \subfloat[$|\widehat{\mathcal{C}}_{\text{II}}|/|\mathcal{C}_{\text{II}}|$ \label{fig:5d}]{
        \includegraphics[width=0.48\textwidth]{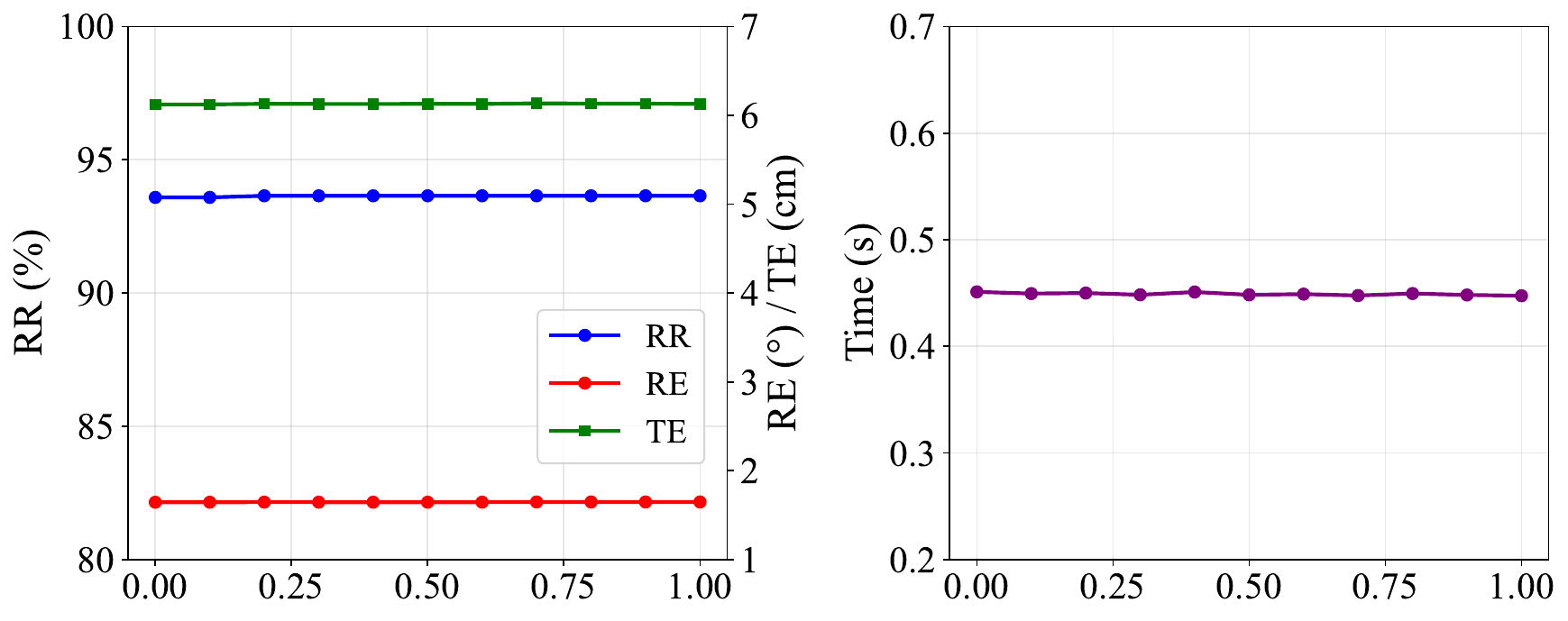}
    }
    \vspace{0.1cm}
    \subfloat[$\lambda$ \label{fig:5e}]{
        \includegraphics[width=0.48\textwidth]{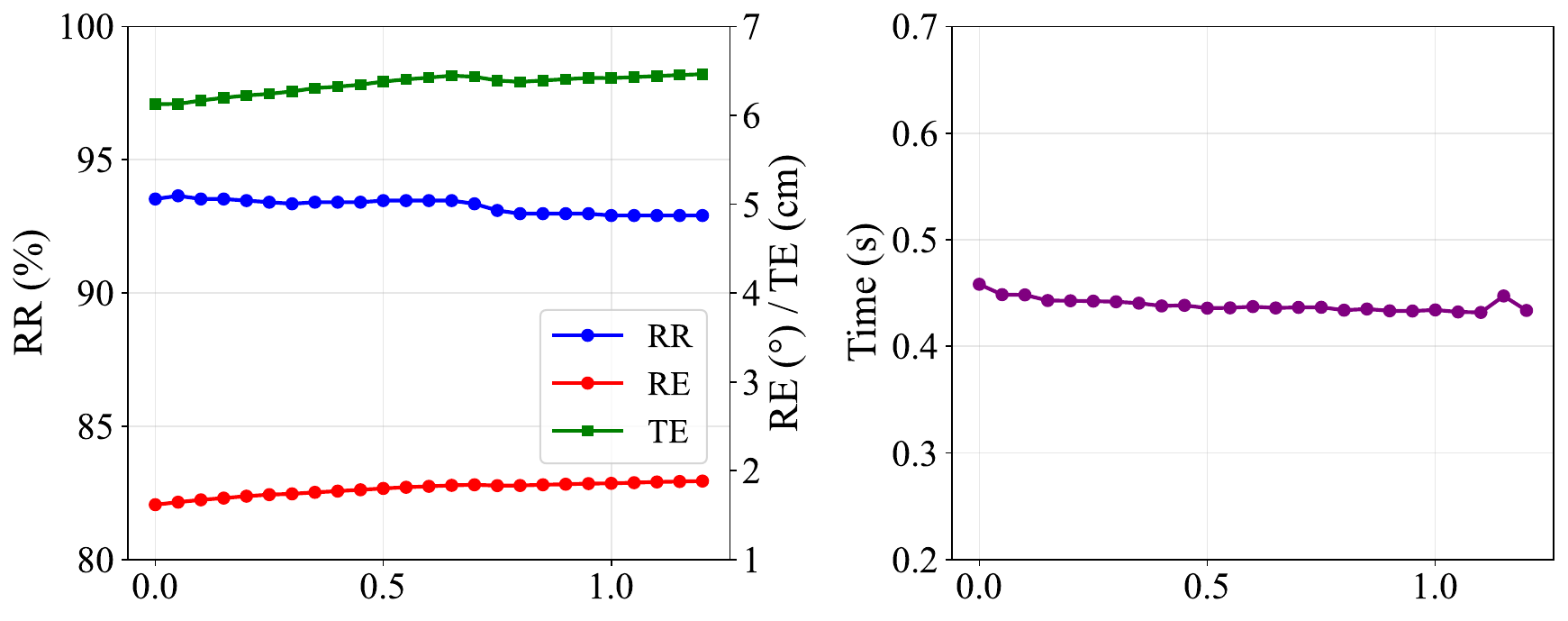}
    }   
    \caption{Parameter sensitivity analysis. The parameter sensitivity analysis curves depict the variation of key performance metrics (RR, RE, TE, Time) with changes in different parameters.}
    \label{fig:parameter}
\end{figure}

\mypara{Parameter $\alpha$}
$\alpha$ is used to determine whether the consensus set found by 1-point RANSAC is sufficiently large to be considered reliable. A higher $\alpha$ value imposes stricter filtering conditions, suitable for low-overlap scenarios (e.g., 3DLoMatch) to filter out noise; whereas a lower $\alpha$ value is suitable for high-overlap scenarios (e.g., 3DMatch) to preserve more true correspondences.

As shown in Figure \ref{fig:5a}, when $\alpha$ varies from 0.0 to 0.8, the Registration Recall (RR) remains stable between 93.6\% and 93.8\%, with minimal fluctuation in error metrics (RE, TE). Performance only slightly degrades when $\alpha$ approaches 1.0. This proves the method's insensitivity to the choice of $\alpha$. The values we set for different datasets based on their overlap characteristics (0.2, 0.9, 0.95) all fall within the stable performance region.

\mypara{Downsampling Parameter}
The input point cloud is downsampled before constructing the geometric proxies to improve the efficiency of subsequent nearest-neighbor search. This parameter primarily affects computational speed, with only secondary effects on accuracy.

Figure \ref{fig:5b} verifies this point. When the downsampling ratio varies from 5$\times$ (our default setting) to 2$\times$ (denser) or 10$\times$ (sparser), the RR consistently remains around 93.5\%. Denser point clouds lead to significantly increased computation time without bringing noticeable improvement in accuracy. This proves that our choice of 5$\times$ resolution is an efficient trade-off between accuracy and efficiency, and the method is insensitive to variations in point cloud density.

\mypara{Parameter $\beta$}
This parameter defines the neighborhood radius for constructing local geometric proxies around anchor points. We set it as a multiple of the point cloud resolution to ensure its adaptability across scenes of different scales. A larger radius captures richer local structures but also increases computational load.

Figure \ref{fig:5c} shows that as the radius $\beta$ increases from 25$\times$ to 60$\times$ the resolution, the RR steadily improves from 93.4\% to 93.7\% and then stabilizes, while RE and TE continuously improve. We set the radius to 50$\times$ the resolution, which is near the "inflection point" where performance saturates and time cost remains acceptable. The results indicate that the method achieves stable and good performance when the radius is sufficiently large (e.g., $>$25$\times$).

\mypara{Parameter $|\widehat{\mathcal{C}}_{\text{II}}|/|\mathcal{C}_{\text{II}}|$}
This parameter is used for robust weight calculation. It determines the width $\sigma$ of the Gaussian kernel function based on the residuals of a high-confidence inlier subset (default: top 40\%). This is an adaptive noise level estimation strategy.

Figure \ref{fig:5d} shows that across the entire range of this parameter from 0.1 to 1.0, all key metrics (RR, RE, TE) remain almost constant, forming nearly horizontal lines. This proves that our robust weight estimation mechanism is very stable.

\mypara{Parameter $\lambda$}
$\lambda$ balances the contribution of feature-space anchors and local geometric-space correspondences in the objective function. For dense indoor point clouds (3DMatch/3DLoMatch), we assign higher weight to geometric information ($\lambda=0.05$); for large-scale, sparse outdoor point clouds (KITTI), we place more trust in the filtered feature anchors ($\lambda=1.0$).

As shown in Figure \ref{fig:5e}, the RR remains above 93.3\% across a wide range of $\lambda$ from 0.0 to 0.5. Although the optimal value is around 0.05, even when $\lambda$ deviates from this value (e.g., increases to 0.5), the performance degradation is minimal. This strongly demonstrates that our dual-space optimization framework effectively integrates both types of information and is insensitive to the specific weight assignment.

The systematic sensitivity analysis demonstrates that our method maintains stable, high performance across a wide range around the default values for all parameters. This confirms the strong robustness of our proposed method.

\end{document}